# Enhancing Data Efficiency and Feature Identification for Lithium-Ion Battery Lifespan Prediction by Deciphering Interpretation of Temporal Patterns and Cyclic Variability Using Attention-Based Models

**Jaewook Lee[a], Seongmin Heo[a], Jay H. Lee[b, *]**

[a]*Department of Chemical and Biomolecular Engineering, Korea Advanced Institute of Science and Technology, 291 Daehak-ro, Yuseong-gu, Daejeon 34141, Republic of Korea*

[b]*Mork Family Department of Chemical Engineering and Materials Science, University of Southern California, 3651 Watt Way, Los Angeles, CA 90089, United States*

[*]*Corresponding author. E-mail address: jlee4140@usc.edu (J.H. Lee).*

## Abstract

Accurately predicting the lifespan of lithium-ion batteries is crucial for optimizing operational strategies and mitigating risks. While numerous studies have aimed at predicting battery lifespan, few have examined the interpretability of their models or how such insights could improve predictions. Addressing this gap, we introduce three innovative models that integrate shallow attention layers into a foundational model from our previous work, which combined elements of recurrent and convolutional neural networks. Utilizing a well-known public dataset, we showcase our methodology's effectiveness. Temporal attention is applied to identify critical timesteps and highlight differences among test cell batches, particularly underscoring the significance of the "rest" phase. Furthermore, by applying cyclic attention via self-attention to context vectors, our approach effectively identifies key cycles, enabling us to strategically decrease the input size for quicker predictions. Employing both single- and multi-head attention mechanisms, we have systematically minimized the required input from 100 to 50 and then to 30 cycles, refining this process based on cyclic attention scores. Our refined model exhibits strong regression capabilities, accurately forecasting the initiation of rapid capacity fade with an average deviation of only 58 cycles by analyzing just the initial 30 cycles of easily accessible input data.

# Acronyms

| | |
|---|---|
| 1D CNN | one-dimensional convolutional neural network |
| 2D CNN | two-dimensional convolutional neural network |
| AM | attention mechanism |
| BMS | battery management system |
| CA | cyclic attention |
| CC | constant current |
| CV | constant voltage |
| CNN | convolutional neural network |
| DHI | direct health index |
| EOL | end of life |
| EN | elastic net |
| EV | electric vehicle |
| GRU | gated recurrent unit |
| RNN | recurrent neural network |
| LAM | loss of active materials |
| LIB | lithium-ion battery |
| LLI | loss of lithium inventory |
| LSTM | long short-term memory |
| NLP | natural language processing |
| MHA | multi-head attention |
| MISO | multi-input single-output |
| OLS | ordinary least squares |
| RHI | refined health index |
| RMSE | root-mean-squared error |
| SA | self-attention |
| SEI | solid electrolyte interphase |
| SHA | single-head attention |
| SOC | state of charge |
| SOH | state of health |
| TA | temporal attention |
| VG | vanishing gradient |
| VIT | voltage, current, temperature |

## Nomenclature

| | |
|---|---|
| $t$ | time step |
| $c$ | cycle |
| $\boldsymbol{ct}$ | context vector |
| $\widetilde{\boldsymbol{ct}}$ | refined context vector |
| $b$ | batch |
| $ce$ | cell |
| $he$ | head |
| $\boldsymbol{AS}$ | attention score matrix |
| $\boldsymbol{HE}$ | head matrix |
| $t_i$ | time step index |
| $c_j$ | cycle index |
| $\boldsymbol{ct}_j$ | context vector for cycle $c_j$ |
| $b_l$ | batch index |
| $ce_m$ | cell index |
| $he_p$ | head index |
| $\boldsymbol{AS}_p, \boldsymbol{HE}_p$ | attention score, head matrix index for $he_n$ |
| $c_{ko}$ | knee-onset |
| $c_{2nd}$ | $2^{nd}$ transition point (cycle) from double Bacon-Watts model |
| $\alpha_1, \alpha_2, \alpha_3$ | estimated model parameters of the double Bacon-Watts model |
| $\gamma, Z$ | adjustable parameter and white noise of the double Bacon-Watts model |
| $V$ | voltage (V) |
| $I$ | current (A) |
| $T$ | temperature (℃) |
| $Q$ | capacity (Ah) |
| $var$ | variable |
| $Q_c$ | charging capacity (Ah) |
| $n_{cy}$ | input cycle size |
| $\boldsymbol{X}$ | input matrix for multi-head attention |
| $cr$ | average C-rate |
| $cr_{1st}$ | C-rate of the 1st charging step |
| $cr_{2nd}$ | C-rate of the 2nd charging step |
| $n_{b_l}$ | number of cells belonging to batch $b_l$ |
| $Q_{tr}$ | point of transition from the 1st to 2nd charging step (Ah) |
| $\boldsymbol{\alpha}_i$ | temporal attention score of $\boldsymbol{h}_i$ |
| $\boldsymbol{h}$ | hidden state vector |
| $\widetilde{\boldsymbol{h}}_i$ | refined hidden state vector |
| $\boldsymbol{w}_b$ | weight vector of temporal attention |
| $n_{layers}$ | number of RNN layers |
| $h_{size}$ | size of each hidden state |
| $\boldsymbol{q}, \boldsymbol{k}, \boldsymbol{v}$ | query, key, value vector encoded from the input vector |
| $\boldsymbol{Q}, \boldsymbol{K}, \boldsymbol{V}$ | query, key, value matrix encoded from input matrix |
| $d_{model}$ | input size |
| $d_q, d_k, d_v$ | query, key, value size |
| $\boldsymbol{w}^Q, \boldsymbol{w}^K, \boldsymbol{w}^V$ | weight vectors for input matrix to embed query, key, value |
| $\boldsymbol{W}^Q, \boldsymbol{W}^K, \boldsymbol{W}^V$ | weight matrices for input matrix to embed query, key, value |
| $he_{size}$ | each head's size |
| $n_{he}$ | number of heads |
| $\boldsymbol{W}^O$ | weight matrix to concatenated heads to single-head size |
| $n_v$ | number of input variables |
| $Q_d$ | discharging capacity |
| $n_{ts}$ | number of time steps per cycle |

# 1. Introduction

As the electric vehicle (EV) market continues to evolve, the importance of the battery management system (BMS) is increasingly prominent. The BMS plays a pivotal role in overseeing the operations of lithium-ion batteries (LIBs) based on real-time measurements [1]. An integral aspect of this system is the accurate prediction of battery lifespan, leveraging the gathered measurements. This predictive capability serves as a valuable tool for users, enabling them to tailor the battery operation to enhance longevity and safety [2]. Furthermore, it offers guidance for achieving optimal battery performance in challenging environmental conditions, such as extremely low and extremely high temperatures.

Conventionally, a LIB is considered to reach its end-of-life (EOL) when the state of health (SOH), defined as the ratio of the current capacity to its nominal capacity, falls to 80% [3]. As the LIB approaches its EOL, the risk of safety accidents, including fire and explosion, escalates [4], underscoring the critical nature of accurate lifespan predictions. However, predicting LIB lifespan proves challenging due to the intricate interplay among various LIB degradation mechanisms, such as Li plating, dendrite formation, electrolyte decomposition, and solid electrolyte interphase formation [5].

## 1.1. Review and evaluation of related literature

The benefits of leveraging data-driven models for understanding physical processes, especially in chemical and biological systems, are increasingly recognized, as evidenced by numerous studies [6-9]. In line with this progression, such models have been adept at forecasting the lifespan of LIBs with high accuracy or estimating their current state (e.g., SOH) without the need for a deep understanding of the underlying mechanisms of LIB degradation [10, 11]. Examples of these models include the elastic net (EN), Gaussian process regression, and support vector regression, all of which have demonstrated effectiveness in this context [12-14]. More recently, neural networks, encompassing a variety of configurations such as convolutional neural networks (CNN) and recurrent neural networks (RNN), have become at the forefront of lifespan prediction methodologies for LIBs [15-18]. There has also been significant interest in combining different machine-learning techniques to create ensemble models, enhancing prediction accuracy and robustness [19, 20]. To better manage the inherent uncertainties in LIB performance, models based on encoder-decoder architectures have been proposed [21, 22]. Moreover, innovative approaches involving CNN-based strategies, including the active utilization of dilated CNNs, have been suggested to improve the estimation of prediction uncertainties [23, 24].

Although existing models for predicting the lifespan of LIBs demonstrate commendable regression capabilities, advancing their interpretability remains a substantial hurdle. It is crucial not only to minimize errors in lifespan prediction for specific datasets but also to understand the rationale behind a model's predictions, including identifying key influential features. An ideal scenario involves leveraging our electrochemical knowledge to validate the significance of these features in the model's performance rather than merely attributing their importance to successful predictions. Such an approach would enhance the model's generalizability and reliability across various datasets, not just the one it was trained on.

To address interpretability concerns, linear models have been popularly employed due to their straightforwardness [12, 25, 26]. However, these models encounter significant drawbacks: 1) they often trade off a considerable degree of predictive accuracy in favor of improved explainability; 2) they offer limited interpretability and regression effectiveness when refined health indicators (RHIs) cannot be directly inputted; and 3) they fail to preserve essential temporal and cyclic features by compressing all input dimensions, thereby overlooking the nuanced characteristics of LIB degradation datasets.

As artificial intelligence continues to make strides across diverse domains [27, 28], its potential to address the limitations of linear models in predicting the lifespan of LIBs is also being realized. The attention mechanism (AM), which originated in the field of natural language processing (NLP), has become a pivotal tool for identifying key and/or negligible elements within datasets [29, 30]. This mechanism has shown its versatility and efficacy in a range of applications, such as image captioning [31], matching images with text [32], and analyzing time series data [33, 34], often being integrated with RNNs to enhance performance [35].

The attention mechanism has also been adapted for use in estimating the SOH and predicting the lifespan of LIBs. While this paper primarily addresses lifespan prediction, it's pertinent to mention SOH estimation techniques, as the AM is predominantly applied in this area. For instance, Yang et al. [36] employed a dual-stage attention mechanism to enhance the estimation of the state of charge (SOC) using long short-term memory (LSTM) [37] and gated recurrent units (GRU) [38]. Zhao et al. [39] utilized the AM to pinpoint key hidden states crucial for SOH estimation. Jiang et al. [40] applied self-attention (SA) within a convolutional autoencoder framework to identify significant features for SOH prediction. Wei et al. [41] optimized a feature graph, including time-related features like 'time to minimum voltage' during discharge, for SOH and remaining useful life (RUL) estimations by incorporating a dual attention mechanism. Lastly, Tang et al. [42] implemented temporal attention (TA) following a combination of CNN and LSTM models to estimate SOH and RUL, integrating RHIs.

Recent studies have also explored the use of the AM to enhance the explainability of models while improving regression performance. For example, Yang [43] integrates 3D CNN, 2D CNN, and feature/cyclic attention mechanisms, attempting to visualize and interpret the significance of each attention score. Wang et al. [44] introduce an adaptive self-attention LSTM model specifically designed to predict the RUL using only charging data. Rahmanian et al. [45] employ an encoder-decoder framework to handle datasets under various experimental conditions, such as cell shape, chemistry, and charging/discharging protocols, aiming to predict metrics like voltage drop, Coulomb efficiency, and discharging capacity. This approach includes an analysis based on saliency maps [46], which evaluate the significance of each input variable by calculating its partial derivative with respect to the output—a post-hoc method employed after training a deep learning model. Lastly, Costa et al. [47] use incremental capacity (IC) curves as input to ascertain the approach to a battery's critical 'knee' point, aiding in the prediction of factors like loss of lithium inventory (LLI), loss of active material (LAM) in the positive electrode and that in the negative electrode. Here, SA is applied to assign varying importance to different segments of the IC curves.

Despite these efforts of recent studies to explore the use of AM in LIB lifespan estimation to both improve regression performance and model explainability, however, certain limitations remain unaddressed as outlined below:

- The current research landscape reveals a gap in connecting high attention scores from models to the underlying electrochemical phenomena or specific operational strategies of LIBs. Notably, there is an absence of qualitative explanations for why certain periods, such as rest phases, might be assigned high TA scores. An example of a missing link is the explanation of how increased internal resistance (IR) during rest periods significantly affects the battery's lifespan. This insight could profoundly deepen our understanding of various degradation patterns across different operational conditions (charging, discharging, resting). Enhancing model generalizability and elucidating the impacts of operational strategies are achievable goals if such connections were made clear. Presently, justifications for attention scores tend to be rudimentary, often correlating high attention to the cycle's end phases merely due to reduced cycle lengths, without associating these findings to specific electrochemical actions or phenomena [43]. This oversight underscores a broader issue in the field, where the potential for leveraging attention scores to provide deeper, actionable insights into battery behavior remains largely untapped.
- There is an identifiable research gap in systematically minimizing the required input data size for accurate lifespan predictions of LIBs, aiming to lower experimental costs by utilizing attention scores. Analyzing outcomes from AM has the potential to significantly reduce the amount of operational cycle data required for precise lifespan predictions, thus facilitating earlier determinations of LIB lifespan. While previous endeavors, such as those by Xiong et al. [48] and our earlier studies [49], have aimed to reduce the amount of input data compared to the methodology of Severson et al. [12] without losing predictive accuracy, a systematic approach for adjusting input size remains undeveloped. In these studies, the determination of input size was often arbitrary or the result of trial and error. Consequently, this lack of a systematic method does not offer a scalable solution for adapting to LIBs with novel chemistries and charge/discharge profiles, thereby limiting the generalizability of these predictive models.

- The complexity of previous models utilizing AM has often hindered the development of model-specific and transparent models. This complexity arises from a focus on enhancing regression performance, making it challenging to understand the training process of the suggested models clearly. While some studies have attempted to improve model explainability through *post-hoc* methods like saliency maps after applying AM to complex structures, these techniques fall short of fully elucidating the training algorithms due to the models' intricacy. Addressing this issue requires a balance between acknowledging the nonlinear nature of complex input-output relationships and striving for simplicity to make the training process transparent without compromising on regression performance. An alternative to deepening the model could involve diversifying the approach within a manageable depth, such as through the use of multi-head attention (MHA). Although MHA has been employed in some studies for predicting the lifespan of LIBs, there is yet to be research demonstrating that MHA can enhance explainability by providing diverse attention score maps while maintaining a low model depth, to our knowledge.

## 1.2. Contributions of this work

In addressing the challenges previously identified, this paper builds upon our earlier work [50], which combined RNN and 1D CNN to predict the lifespan of LIBs. We have enhanced this approach by incorporating TA and cyclic attention (CA), leading to the creation of three separate models: RNN + TA + 1D CNN, RNN + CA + 1D CNN, and RNN + TA + CA + 1D CNN. These models are designed to improve upon the regression performance of LIB lifespan predictions while also providing the ability to identify key timesteps and cycles through the interpretation of TA and CA scores. By maintaining a simple model architecture, we aim to reduce computational costs and improve the ease with which AM-based models can be interpreted, mirroring the way in which regression models are trained to distill crucial information. Recognizing the significance of both intra-cycle and inter-cycle dynamics in lifespan prediction, we have structured the timestep dimension to distinguish between cycles and timesteps within cycles, facilitating the extraction of relevant behaviors through RNN and 1D CNN. To further the models' generalizability with limited data, we utilized three subsets from the main dataset [12] (i.e., the *Severson dataset*): charging only, discharging only, and a combined dataset that includes rest phases, considering the necessity to adapt to various charge and rest profiles for accurate lifespan prediction. The contributions of this paper that advance the field beyond the current state of existing literature are outlined as follows:

- By incorporating a simple TA layer into the hidden states of RNN, we can pinpoint critical timesteps with precision. This method calculates the TA scores for each hidden state directly, facilitating the identification of crucial timesteps without adding substantial complexity to the model. To demonstrate the effectiveness of the calculated TA scores, we highlight batch-to-batch variations within the Severson dataset, particularly noting significant differences in rest period lengths between test cell batches, which correlate to considerable disparities in lifespan. Additionally, applying TA across hidden states to form a context vector addresses issues such as vanishing gradients (VG) and loss of information from earlier inputs as the number of timesteps increases by considering all hidden states simultaneously through a single context vector for each cycle.
- To systematically reduce the required input size and consequently lower experimental costs for LIB lifespan predictions, we employed SA on context vectors, each representing a cycle. Analyzing SA scores through key-query relationships enables the identification of pivotal cycles that demonstrate high correlations with all query cycles, suggesting these key cycles contain the bulk of information across the dataset. This input size reduction technique, informed by CA scores, offers a straightforward and practical approach for manufacturers to implement with their data, adding just a single layer of depth to the baseline model. Our experimental analysis revealed that cycles 20-40 were particularly correlated to all query cycles, facilitating a reduction in the input size from 100 to 50 cycles without detracting from the model's regression accuracy.

- To consider complex input-output relationships from multiple angles and refine the input reduction process, MHA is extensively employed while keeping the model depth comparable to SHA. By diversely training the query, key, and value parameters, researchers can identify distinct crucial key cycles for each query. The number of attention heads is increased until no new patterns in CA scores are discovered, aiding in the systematic determination of the number of cycles to reduce input size. In our experiments, optimizing to three heads identified cycles 0-20 as essential, enabling a reduction in input size to 30 cycles from an initial 100, with a reasonable regression loss on the test set.
- An additional key aspect of our work is the exclusive use of direct health indicators (DHIs), such as voltage ($V$), current ($I$), temperature ($T$), and capacity during charging ($Q_c$) or discharging ($Q_d$), across charging only, discharging only, and combined datasets. This approach allows battery manufacturers to employ the proposed models in real-time without the need for post-processing or conducting additional refined reference performance tests to calculate refined health indicators (RHIs, e.g., IR [12]). The final model, RNN + TA + CA + 1D CNN, demonstrated superior performance, achieving an error margin of only 55-60 cycles compared to models that utilize RHIs as input [12].

## 1.3. Organization of this paper

The remainder of this paper is structured as follows: Chapter 2 offers a concise overview of the Severson dataset, including input/output formulations and preprocessing methodologies, elaborating on details previously introduced in our work [50]. Chapter 3 revisits baseline models from our earlier research [50], and proceeds to elaborate on the application of TA, SA, and MHA within the context of LIB lifespan prediction. This chapter also provides an in-depth discussion of the three proposed models that integrate AM with RNN + 1D CNN frameworks, showcasing how attention scores facilitate the identification of crucial timesteps under varied experimental conditions and enable a systematic approach to input data size reduction. Chapter 4 presents the experimental outcomes, beginning with a scenario utilizing 100 cycles of input data to juxtapose the proposed models against the baselines.

Following this, an analysis of TA and CA scores is performed. Subsequently, in a second scenario, input size reduction experiments leveraging CA scores for SHA and MHA are executed and evaluated. The paper concludes with Chapter 5, summarizing the findings and contributions.

# 2. Problem formulation

## 2.1. Input and output formation

The primary goal of this study is to devise a machine-learning technique that efficiently utilizes readily available information from a relatively small number of initial charge and/or discharge cycles of LIBs to predict their lifespan. Our specific aim is to predict *knee-onset*, which signifies the initiation of nonlinear capacity degradation, utilizing data from a fixed number of initial cycles. Knee-onset can be estimated from the capacity degradation curve with the double Bacon-Watts model outlined below:

$$Q = \alpha_0 + \alpha_1(c - c_{ko}) + \alpha_2(c - c_{ko})\tanh\{(c - c_{ko})/\gamma\} + \alpha_3(c - c_{2nd})\tanh\{(c - c_{2nd})/\gamma\} + Z \quad (1)$$

where $Q$ represents the capacity, $c$ denotes the cycle number, $c_{ko}$ is the knee-onset, $c_{2nd}$ is the second transition point, $\alpha_1, \alpha_2$ and $\alpha_3$ are the model parameters to be estimated, $\gamma$ is an adjustable parameter, and $Z$ is a white noise [51]. It is well-established that IR increases significantly after the knee-onset, leading to nonlinear degradation behavior [52]. The knee-onset of some cells is illustrated in Fig. S1, and detailed values for each cell are provided in Table S1.

The input variables for the mainly utilized dataset include $V$, $I$, $T$, $Q_c$ and $Q_d$. For a variable $var$, $var_{c_j,t_i}^{ce_m}$ represents its value at time $t_i$ of cycle $c_j$ (cell $ce_m$). Let $n_{cy}$ and $n_{ts}$ be the number of input cycles used and the total number of time steps per cycle, respectively. The input data matrix $\boldsymbol{X}^{ce_m}$ can be formed for cell $ce_m$ can then be formulated as follows:

$$\boldsymbol{X}^{ce_m} = \begin{bmatrix} V_{1,1}^{ce_m} & \cdots & V_{1,n_{ts}}^{ce_m} & \cdots & \cdots V_{n_{cy},1}^{ce_m} & \cdots & V_{n_{cy},n_{ts}}^{ce_m} \\ I_{1,1}^{ce_m} & \cdots & I_{1,n_{ts}}^{ce_m} & \cdots & \cdots I_{n_{cy},1}^{ce_m} & \cdots & I_{n_{cy},n_{ts}}^{ce_m} \\ T_{1,1}^{ce_m} & \cdots & T_{1,n_{ts}}^{ce_m} & \cdots & \cdots T_{n_{cy},1}^{ce_m} & \cdots & T_{n_{cy},n_{ts}}^{ce_m} \\ Q_{c_{1,1}}^{ce_m} & \cdots & Q_{c_{1,n_{ts}}}^{ce_m} & \cdots & \cdots Q_{c_{n_{cy},1}}^{ce_m} & \cdots & Q_{c_{n_{cy},n_{ts}}}^{ce_m} \\ Q_{d_{1,1}}^{ce_m} & \cdots & Q_{d_{1,n_{ts}}}^{ce_m} & \cdots & \cdots Q_{d_{n_{cy},1}}^{ce_m} & \cdots & Q_{d_{n_{cy},n_{ts}}}^{ce_m} \end{bmatrix} \quad (2)$$

## 2.2. Dataset description

In this part, we describe the Severson dataset and some steps we performed to characterize and preprocess it to construct $\boldsymbol{X}^{ce_m}$ properly. Since detailed explanations of the whole dataset were given in [50], we focus on the differences among the three different batches operating protocols and their effect on the knee-onset. Among the several available public datasets for LIB degradation [53, 54], the Severson dataset is chosen as it is a relatively large dataset involving 124 normal cells. This dataset was collected under a well-controlled setup, keeping the environmental variables (e.g., ambient temperature) constant. Therefore, it has become a popular benchmark dataset to test data-driven modeling approaches in the literature [55, 56].

Let us briefly discuss the main features of the dataset. Each cell was subjected to repeated tests of charging-discharging cycles until its EOL. During the charging phase, one or two-step constant current (CC) policies were applied up to 80% SOC. Then, a certain amount of rest time was placed (details will be discussed shortly). After the rest, a constant voltage (CV) charging policy was applied to reach 100% SOC. The discharging phase employed the same 4C CC policy for all the cells. Finally, the discharging was followed by another rest time to complete the cycle. The dataset encompasses three batches of data, which feature different rest time characteristics. In Batch 1, different rest times were applied after 80% SOC charging (1 min) and after discharging (1 sec), while the same rest time was used for both rests in Batch 2 (5 min) and Batch 3 (5 sec). In Batch 3, two additional rests of 5 sec were applied after the IR test and before the discharging.

Fig. S2(a) shows each batch's average C-rate until 80% SOC (end of CC policy) and IR values (averaged only for the last 100 cycles of each cell). The average C-rate until 80% SOC of batch $b_l$, $cr_{b_l}$, can be calculated using the following Eq. (3).

$$cr_{b_l} = \frac{\sum_{m=1}^{n_{b_l}} \left( \frac{cr_{1st}^{ce_m} Q_{tr}^{ce_m}}{80} + \frac{cr_{2nd}^{ce_m} \left(80 - Q_{tr}^{ce_m}\right)}{80} \right)}{n_{b_l}} \tag{3}$$

where $n_{b_l}$ is the number of cells belonging to batch $b_l$, $Q_{tr}^{ce_m}$(%) is the point of transition from the 1st to 2nd charging step for cell $ce_m$, and $cr_{1st}^{ce_m}$ and $cr_{2nd}^{ce_m}$ are the C-rates of the 1st charging step (0-$Q_{tr}^{ce_m}$% SOC) and the 2nd charging step ($Q_{tr}^{ce_m}$-80% SOC) for cell $ce_m$ of batch $b_l$, respectively.

Looking at $cr_{b_l}$ of each batch, although different charging policies were applied (72 policies in total), the average charging rate varies by very little across the batches. We can also see that Batch 2 has the largest IR value, followed by Batch 1 and Batch 3, exactly matching the order of the rest times. In Keil et al. [57], it was shown that an increase in rest time led to a considerable increase in IR and such an increase is accelerated when the SOC is large. As mentioned above, in the Severson dataset, the rest occurs once at 80% SOC, leading to a significant increase in IR (especially for Batch 2).

Fig. S2(b) summarizes the average and standard deviation of knee-onset for each batch. In Celik et al. [58], it was analyzed that the Spearman coefficient between the IR and the cycle life in the Severson dataset was -0.6517, implying a strong correlation between them, which can be confirmed by inspecting Fig. S2(a), (b). Looking at the standard deviations of the knee-onset, larger variations can be observed for Batch 1 and Batch 3 compared to Batch 2. This is mainly due to the relatively short lifespans of the cells in Batch 2.

Based on the above, it can be argued that it is important for a knee-onset regression model to achieve small errors for Batches 1 and 3, which exhibit large variations in the lifespan. Rest time, a design variable for cycling tests (thus readily available), has the potential to be used as an indicator of IR, which is expensive to measure. These ideas will be tested using the data-driven models described subsequently.

In our test, the number of initial cycles in the input data to predict knee onset will vary. Specifically, $n_{cy}$ takes different values (30, 50, 80, and 100), with 100 being the base case value. Within each cycle, samples are taken with a 0.5 min interval up to 60 min (i.e., $n_{ts} = 120$), and zero-padding is used when needed to accommodate variable cycle lengths. When using the charging-only dataset with only the measurements taken during charging operations, we use $V, I, T$ and $Q_c$ as the inputs with $n_{ts} = 40$. On the other hand, when using the discharging-only dataset, $dQ/dV, Q_d, T$ are used as the inputs. The input dataset is rearranged with respect to voltage (as a surrogate of time) since only constant-current policy is used in the discharging phase, and $n_{ts}$ is selected as 1000 points. The reasons for these choices were explained in our previous research [50].

## 2.3. Preprocessing, data split, and main hyperparameters

In this part, preprocessing methods are described briefly before introducing developed algorithms in this work in Chapter 3. The Severson dataset contains some outliers, which are removed and replaced by extrapolated values adhering to the approach described by Yang [43], and the noise level is reduced using the Savitzky-Golay filter [59]. Subsequently, min-max normalization is applied to scale the input values from 0 to 1. To ensure representative results (independent of random grouping), five distinct random selections of training, validation, and test sets are attempted, and their results are averaged.

For the dataset split, we randomly selected 80, 20, and 24 cells among the 124 cells, designating the corresponding data sets as training, validation, and test sets. To assess the impact of initialization, we employed five different splits of training, validation, and test sets generated by different random seeds. We calculated the average and standard deviation of the test losses. The error bars of test loss, computed from the results of five different seeds, are presented in Fig. S6~S8, demonstrating the consistency of the model's performance concerning the initialization.

Based on the outcomes of preliminary tests involving multiple RNN types, GRU was selected as the RNN type. To counteract overfitting, early stopping was implemented, where [epoch, patience] was set to [3000, 500] for training using data from all the batches and [1000, 10] for training with data from each batch, respectively. Regarding MHA, the number of heads varied from 1 to 5, and the detailed hyperparameter values we settled on are provided in Table S6~S10.

# 3. Methods

## 3.1. Baseline models: RNN + 1D CNN, RNN + 2D CNN

Fig. 1 illustrates the data-driven models developed in our previous research [50], which will serve as the starting points for the models developed in this study. Two constituents of these models are CNN and RNN. CNN, a neural network type designed for extracting spatially local features through convolution kernels [60], is combined with RNN, widely used for time series modeling, where the outputs of the hidden states serve as inputs to the same nodes for subsequent time step calculations.

In the case of CNN, the limited span of the kernel used [61] necessitates many layers to effectively extract information regarding the entire cycle, potentially causing the VG problem. In our earlier work, we proposed two data-driven models, namely RNN + 1D CNN and RNN + 2D CNN [50], by integrating RNN with CNN. In RNN + 1D CNN, the last hidden states of the RNN are used as inputs to the 1D CNN. Incorporating RNN to encode input information reduced the layer depth of the CNN and training time. However, the VG problem persisted as the time series of a cycle could be a long sequence.

In RNN + 2D CNN, the entire sequence of hidden states from the RNN is used as inputs to the 2D CNN. This model helped mitigate the VG problem and enhance accuracy, albeit with a longer training time. The major drawback of RNN + 2D CNN lies in its high memory requirements and training difficulty, which can become pronounced with increasing data sampling rates or decreasing charging rates.

In addition, accurately predicting the battery's lifespan and understanding the underlying reasons for such predictions are crucial. The two models that integrate RNN and CNN architectures face the limitation of not providing direct explanations about which specific timesteps or cycles are most influential in determining the lifespan. The attributes of both the baseline and the proposed models are concisely outlined in Table 1.

To address the limitations of the baseline models highlighted previously, this study incorporates three AM-based frameworks into the baseline model. These enhancements not only overcome the initial shortcomings but also amplify the model's explanatory capabilities, streamline the input size, and elucidate batch-to-batch variations within the input dataset.

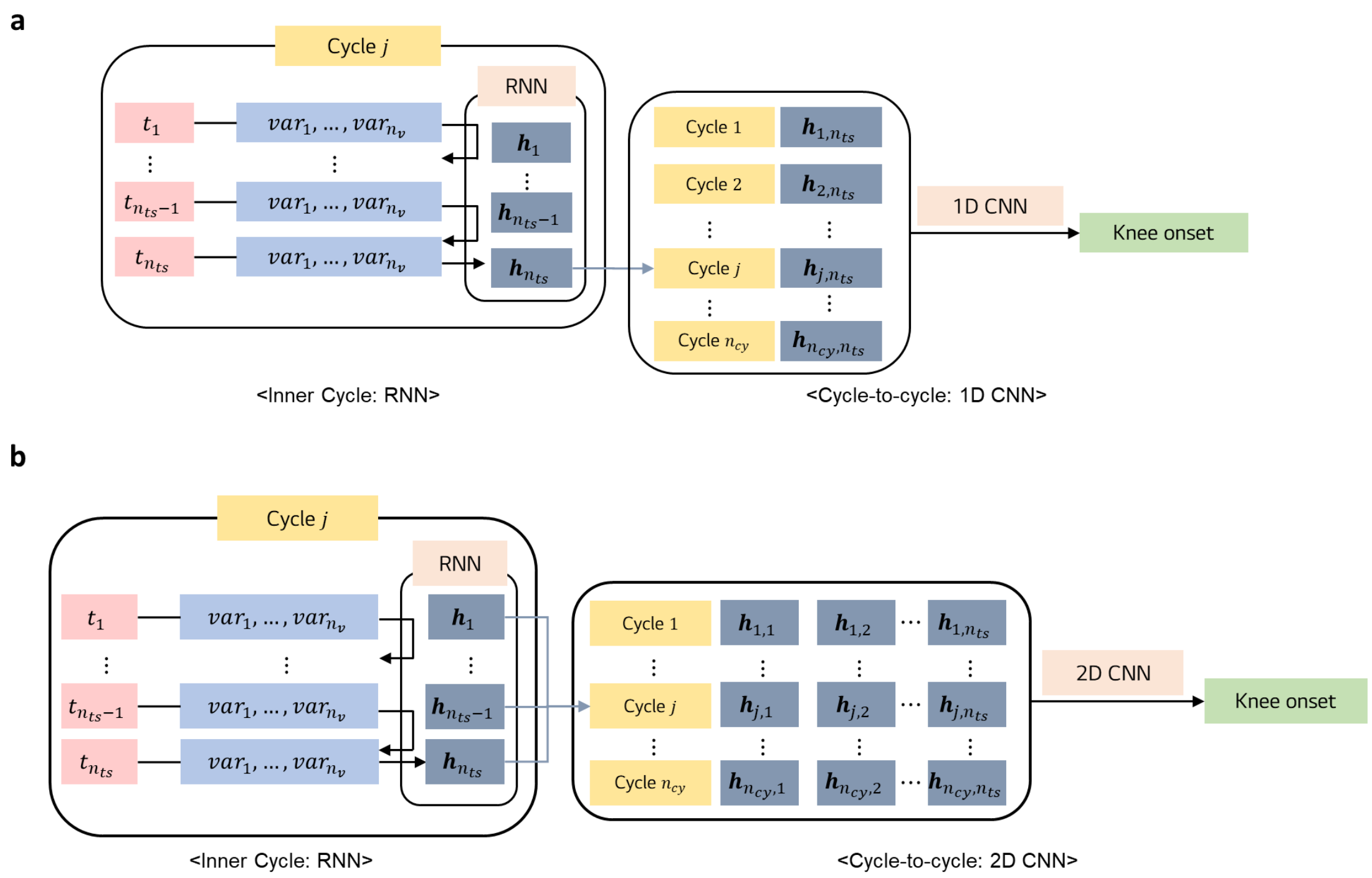

Fig. 1. Baseline models from our previous research [50]. (a) RNN + 1D CNN model, (b) RNN + 2D CNN model.

## 3.2. Attention mechanisms employed in this study

Before describing the actual models utilized in this study, we would like to introduce the attention frameworks utilized in this work. Three different frameworks (i.e., temporal attention (TA), self-attention (SA), and multi-head attention (MHA) were utilized, and we will introduce the features of each framework and how they can be applied to time series and cyclic datasets for LIB’s lifespan prediction.

### 3.2.1. Temporal attention

TA is an attention-based algorithm commonly applied to sequential data, such as those in NLP and time-series modeling. TA functions by assigning varying weights to previous inputs based on their influence in generating the output of a desired sequence. Using these weighted inputs, a *context vector* is calculated, flexibly reflecting the impact of previous inputs. The TA score for each element of the sequence indicates its importance. It is worth noting that although it is possible to use non-RNN models, such as feedforward neural networks, for sequential data, this is beyond the scope of our study.

When TA is integrated with RNNs, the hidden states serve as inputs to the TA layer. We focus on TA for multi-input-single-output (MISO) cases as the output has a scalar value (knee-onset in our case). The two main equations used to calculate the TA score of a hidden state of time step $t_i$ in an RNN-based MISO problem are shown below.

$$\boldsymbol{\alpha}_i = const. \tag{4}$$

$$\boldsymbol{\alpha}_i = \frac{\exp\left(\boldsymbol{w}_b \boldsymbol{h}_i^T\right)}{\sum_{i=1}^{n_{ts}} \exp\left(\boldsymbol{w}_b \boldsymbol{h}_i^T\right)} \tag{5}$$

Eq. (4) fixes the attention score at each time step $t_i$ to a constant value, which is simple and meaningful if the relationship between time steps is consistent throughout the input data. However, it is unsuitable in our case due to different cells employing distinct charging and resting strategies, leading to dynamically changing relative importance of time steps.

Eq. (5) keeps the weight vector $\boldsymbol{w}_b$ fixed during training but adjusts a TA score based on the incoming hidden state. Since the importance of a time step can be set differently depending on the value of the hidden state, it allows for variations in the importance of time steps in dynamic data. Note that, as the hidden state is also a result of the learned RNN layer, training should be conducted to ensure that the calculated attention output accurately reflects the importance. Details of TA with dynamic attention scores are depicted in Fig. 2. After calculating the refined hidden state $\widetilde{\boldsymbol{h}}_i = \boldsymbol{w}_b \boldsymbol{h}_i^T$ from each hidden state $\boldsymbol{h}_i$, the TA score from each time step is calculated from the refined hidden state and multiplied to each hidden state.

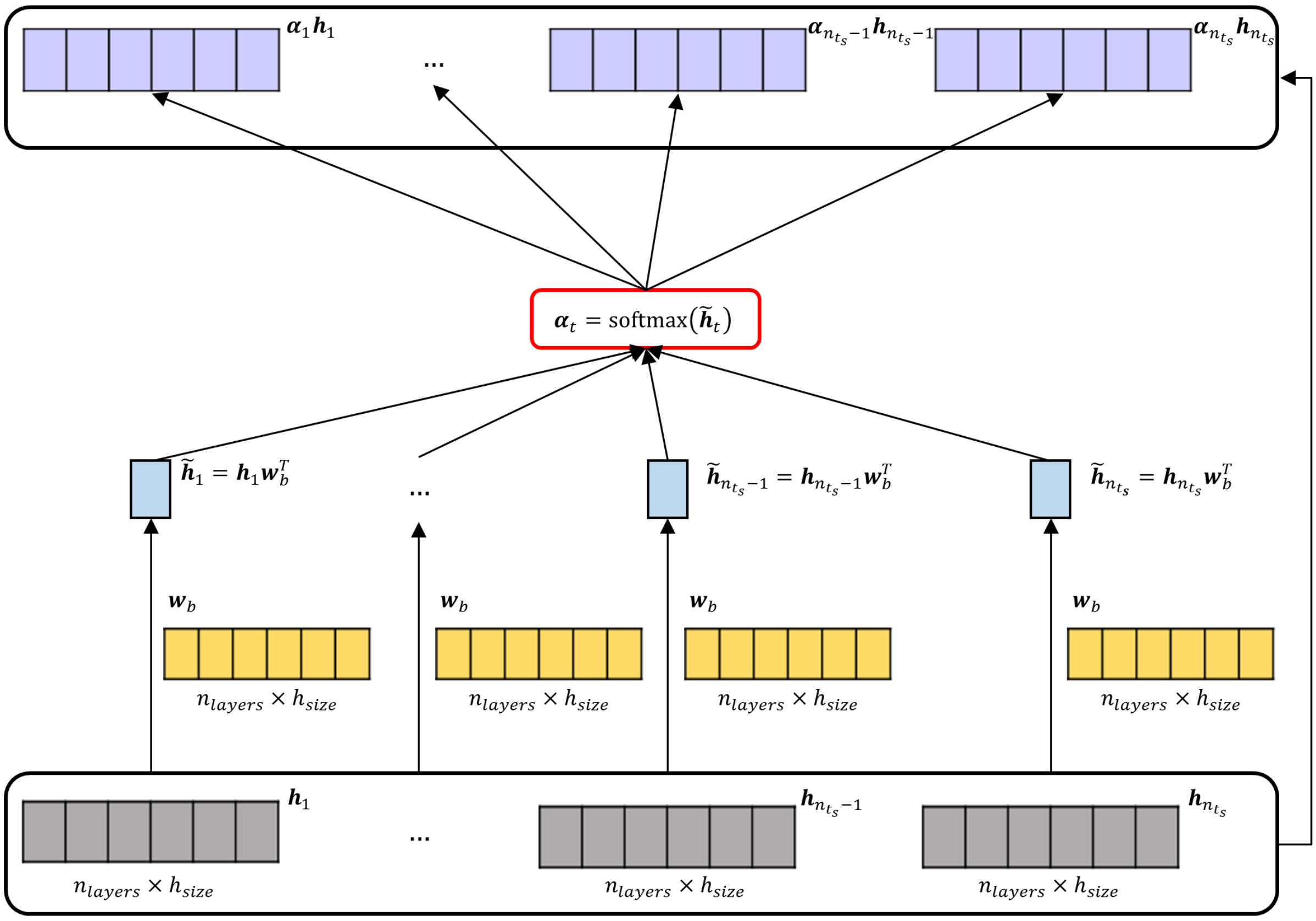

Fig. 2. Schematic of temporal attention applied in the task of predicting the lifespan of LIBs. For each hidden vector, weight matrix $\boldsymbol{w}_b$ is multiplied, then the TA score of each hidden state, $\boldsymbol{\alpha}_t$ is calculated with softmax function. Subsequently, these TA scores are multiplied by the original hidden states to emphasize the significance of crucial time steps. The terminology used includes $n_{layers}$, which denotes the number of RNN layers, and $h_{size}$, which indicates the size of each hidden state as represented by Eq. (5). In this approach, we opted for $n_{layer} = 1$ to maintain minimal model depth.

### 3.2.2. Self-Attention

SA is a recently proposed attention mechanism in Transformer architectures [30], and its application to various NLP problems has garnered significant attention, as demonstrated in Devlin et al. [62]. In the Transformer framework, SA estimates relationships between different positions within a given sequence. When used with RNN, SA utilizes the similarity among hidden states to give higher weights to more representative hidden states. SA introduces three concepts to calculate the attention score: *query, key*, and *value*. Further details about each concept can be found in Niu et al. [63].

SA can extract key representative features of cycle-to-cycle behavior for cyclic datasets based on cycles' similarities. In the knee-onset prediction problem, this can aid in reducing the number of input cycles: if a specific cycle already possesses much information from other future cycles in the input, then the input to predict knee-onset can be effectively reduced to that specific cycle. In our problem, inputs for the SA are computed from the hidden states of each cycle, referred to as the context vectors (denoted as $\boldsymbol{ct}$ for the sake of our discussion below).

Let the current cycle be $c_j$ among $n_{cy}$ cycles. The similarity between the context vectors from the RNN at $c_j$, $\boldsymbol{ct}_j$ and that at another cycle $c_{j'}$, $\boldsymbol{ct}_{j'}$ can be measured by SA. In this case, query ($\boldsymbol{q}$) is the vector encoding $\boldsymbol{ct}_j$ and key ($\boldsymbol{k}$) is the vector encoding $\boldsymbol{ct}_{j'}$. Value ($\boldsymbol{v}$) is a vector representing the actual context vector associated with the key, i.e., $\boldsymbol{ct}_{j'}$. Additionally, let the size of the input originally belonging to each sequence be $d_{model}$, and the size of the query, key, and value $d_q, d_k, d_v$. In SA, $d_q, d_k, d_v$ are all equal.

We multiply the encoded query and key to calculate the similarity between the two targets, which is the SA score. As the size of the key increases, the similarity between each variable weakens. To counteract this, we divide it by $\sqrt{d_k}$ and apply the softmax function to normalize it to a value between 0 and 1. In this context, $\text{softmax}\left(\frac{\boldsymbol{q}\boldsymbol{k}^T}{\sqrt{d_k}}\right)$ can be called the SA score for cycles $c_j$ and $c_{j'}$. Finally, the value vector is multiplied by the SA score to compute the actual output of each target associated with the query, which is the final output of the SA. If we extend this calculation to the entire input matrix X and denote the query, key, and value of X as $\boldsymbol{Q}, \boldsymbol{K}, \boldsymbol{V}$, we can calculate the SA score matrix $\boldsymbol{AS}$, as shown in Eq. (7). A detailed schematic of the SA is presented in Fig. 3. For the output, the head of the SA layer is denoted as $\boldsymbol{HE}$.

Building on the key-query concept, we can attempt to reduce the input data size in predicting knee-onset as follows. The SA score of each query cycle is calculated for each key cycle. Then, suppose the SA score of a specific key cycle is consistently high across all query cycles. In that case, it indicates that the context vector of the key cycle already encapsulates crucial information representative of all the cycles. Consequently, the regression model can be constructed with only those cycles identified as important, presenting the potential to decrease the number of cycles needed to predict knee-onset.

$$\boldsymbol{Q} = \boldsymbol{X}\boldsymbol{W}^{Q}, \boldsymbol{K} = \boldsymbol{X}\boldsymbol{W}^{K}, \boldsymbol{V} = \boldsymbol{X}\boldsymbol{W}^{V} \tag{6}$$

$$\boldsymbol{HE} = \text{SelfAttention}(\boldsymbol{Q}, \boldsymbol{K}, \boldsymbol{V}) = \text{softmax}\left(\frac{\boldsymbol{Q}\boldsymbol{K}^{T}}{\sqrt{d_k}}\right)\boldsymbol{V} = \boldsymbol{ASV} \tag{7}$$

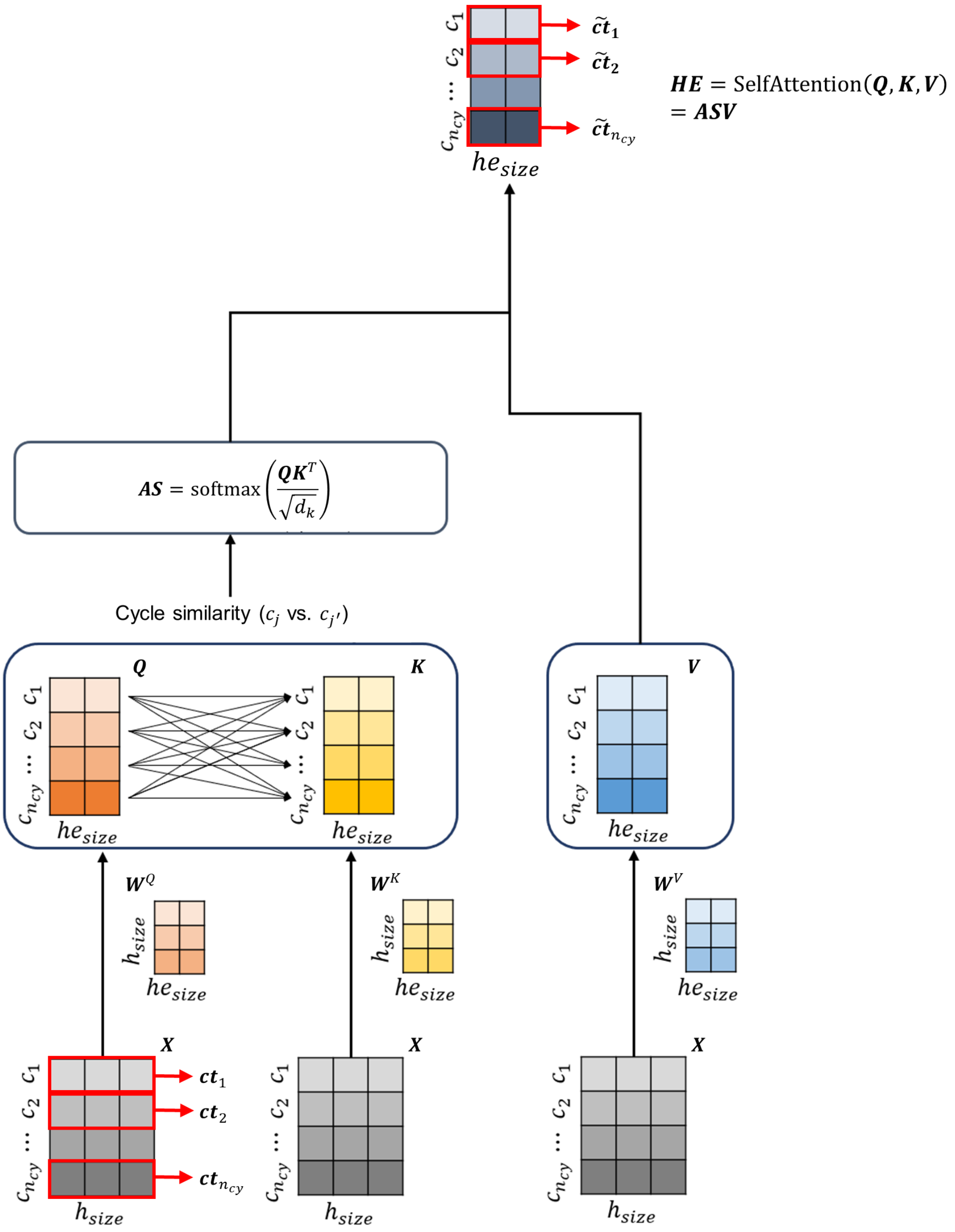

Fig. 3. Schematic of self-attention (SA) applied to context vectors, following their calculations from an RNN (+ TA). $he_{size}$ refers to the size of the head in the self-attention layer. For context vector matrix $\boldsymbol{X}$,

query, key, and value matrices are calculated, then the attention score $(\mathbf{AS})$ is calculated using the formula $\mathbf{AS} = \text{softmax}\left(\frac{\boldsymbol{QK}^T}{\sqrt{d_k}}\right)$ where $d_k$ is the dimensionality of the key. This attention score is then applied to the value matrix to generate the SA output.

### 3.2.3. Multi-head attention

A key advantage of CNN lies in its ability to utilize multiple kernels, enabling the examination of the same input from various perspectives. Different strides, kernel sizes, and dilations can be employed to explore the input data thoroughly, and the diverse information acquired from these different kernels can be concatenated to form a highly relevant output.

MHA seeks to replicate the functionality of multiple feature maps in CNN by investigating input sequences from multiple viewpoints [30]. Although the receptive field of attention remains consistent across all inputs, MHA assigns distinct attention scores for different heads. Instead of designating a particular input as crucial, MHA allows the neural network to use diverse perspectives to assess the importance of each input. Like the multiple kernels in CNN, MHA has demonstrated its capability to enhance accuracy and stability compared to SHA.

In the knee-onset prediction problem, where inputs and outputs exhibit complex relationships, MHA could prove beneficial, as SHA may overlook some critical connections between inputs and outputs. With MHA, different input parts would be analyzed in diverse ways, reducing the likelihood of omitting essential information.

The mathematical representations of MHA are as follows. Although different inputs can be query and key/value for MHA, in our case, where MHA is combined with SA, all inputs for query, key, and value are denoted as $\boldsymbol{X} \in \boldsymbol{R}^{n_{cy} \times d_{model}}$ ($d_{model} = h_{size}$). Moreover, the sizes of query, key, and value are set equal in this approach, specifically $d_q = d_k = d_v = he_{size}$. The projection matrices for the $p$-th head are denoted as $\boldsymbol{W}_p^Q \in \boldsymbol{R}^{d_{model} \times d_q}$, $\boldsymbol{W}_p^K \in \boldsymbol{R}^{d_{model} \times d_k}$, and $\boldsymbol{W}_p^V \in \boldsymbol{R}^{d_{model} \times d_v}$. Consequently, the query, key, and value matrices become $\boldsymbol{Q}_p = \boldsymbol{X}\boldsymbol{W}_p^Q, \boldsymbol{K}_p = \boldsymbol{X}\boldsymbol{W}_p^K, \boldsymbol{V}_p = \boldsymbol{X}\boldsymbol{W}_p^V$. After calculating the attention score $\boldsymbol{AS}_p$ for each head using $\text{softmax}\left(\frac{\boldsymbol{Q}_p\boldsymbol{K}_p}{\sqrt{d_k}}\right)$, the output of each head $\boldsymbol{HE}_p$ is calculated by multiplying $\boldsymbol{AS}_p$ and $\boldsymbol{V}_p$ as Eq. (8). Then, MHA can be performed by concatenating all the heads and multiplying the resulting matrix with $\boldsymbol{W}^O \in \boldsymbol{R}^{d_v \times n_{he} d_v}$, as shown in Eq. (9).

$$\boldsymbol{HE}_p = \text{SelfAttention}(\boldsymbol{Q}_p, \boldsymbol{K}_p, \boldsymbol{V}_p) = \text{softmax}\left(\frac{\boldsymbol{Q}_p \boldsymbol{K}_p^T}{\sqrt{d_k}}\right)\boldsymbol{V} = \boldsymbol{AS}_p \boldsymbol{V}_p \tag{8}$$

$$\text{MultiHead}(\boldsymbol{Q}, \boldsymbol{K}, \boldsymbol{V}) = \text{Concat}(\boldsymbol{HE}_1, \dots, \boldsymbol{HE}_{n_{he}})(\boldsymbol{W}^O)^T \tag{9}$$

Different weight matrices for each head are assigned to queries, keys, and values. The output of each head is then concatenated. To manage the output size, the matrix $\boldsymbol{W}^O$ in the last step regulates the size as a simple concatenation of multiple heads' outputs could result in a large output size.

The Transformer architecture demonstrated that various heads produced distinct attention scores, allowing them to interpret complex sentences in multiple ways to uncover meaningful relationships. In our study, MHA is used with SA to identify important cycles, enabling the analysis of significant cycles from various perspectives based on similarities between context vectors of different cycles. If the regression model is appropriately trained, different heads capture diverse relationships between context vectors, thereby giving a more nuanced representation of the dataset. We analyzed the learned attention scores from each head to enhance regression performance and interpretability. Detailed steps of MHA are illustrated in Fig. 4.

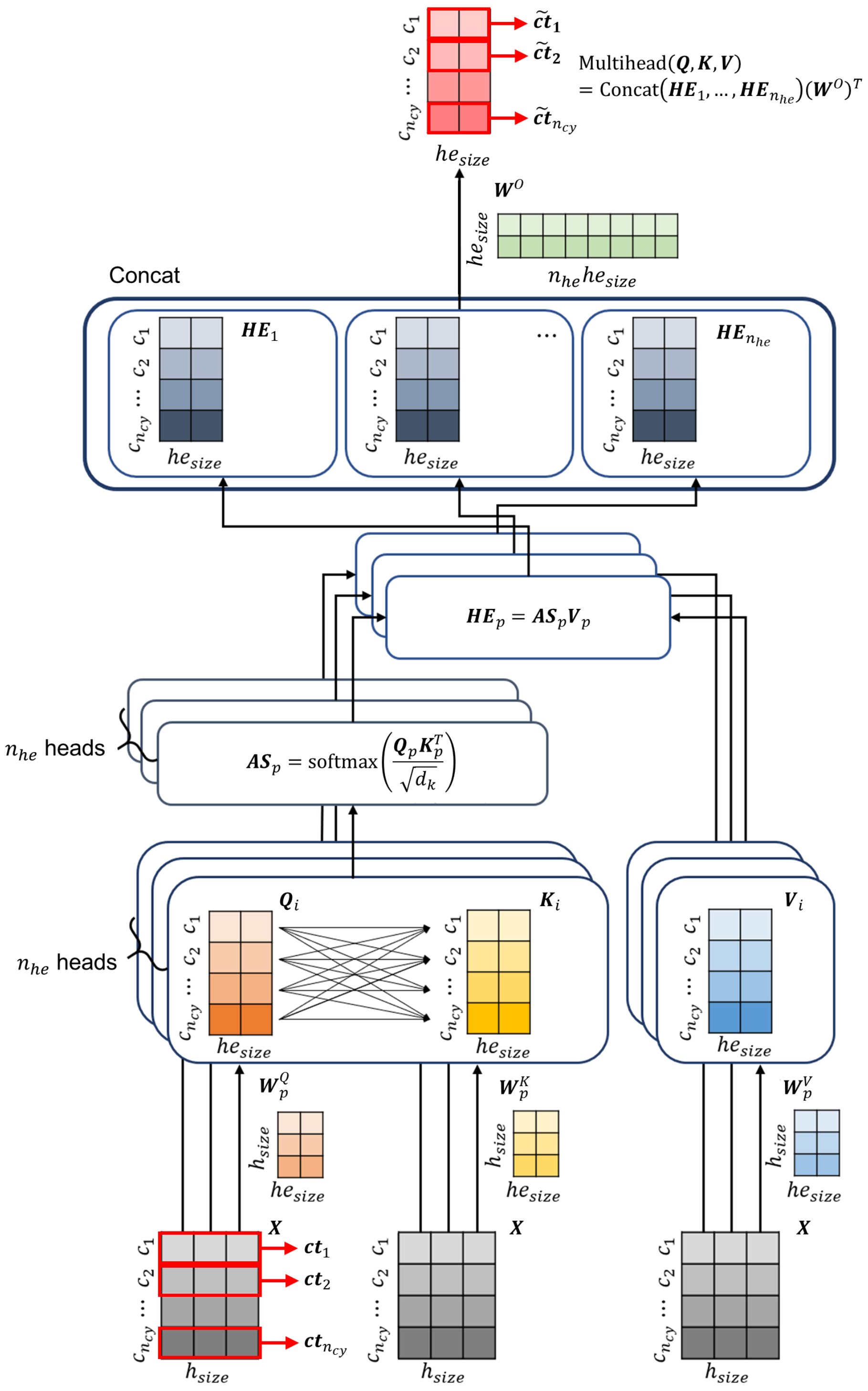

Fig. 4. Schematic of multi-head attention as applied in the task of predicting the lifespan of LIBs. For this application, SA principles are employed, meaning that the inputs for the query, key, and value matrices are all equal to $\boldsymbol{X}$. Similar to the process detailed in Fig. 3, for each head within the multi-head framework, the query, key, and value matrices are computed. Subsequently, the outputs from each head are concatenated and then projected down to match the dimensionality of a single head, thereby

standardizing the SA output regardless of the number of heads employed. $n_{he}$ refers to the number of heads in the MHA framework.

## 3.3. Proposed models

With suggested attention frameworks, we suggest several knee-onset regression models for LIBs that encapsulate the crucial characteristics of intra- and inter-cycle behaviors inherent in LIB degradation data. These models leverage TA, SA, and MHA for baseline models [50].

### 3.3.1. RNN + TA + 1D CNN

In this work, TA is used to gauge correlations between different time steps within the same cycle to capture relevant features of intra-cycle trends, which entail dynamic variations within each cycle. TA mitigates the VG problem by using the entire hidden state sequence to calculate the context vector, serving as input to the 1D CNN. Notably, the RNN + 1D CNN model without TA encounters the VG problem, relying solely on the last hidden state of each cycle as the context vector.

Fig. 5 illustrates the proposed model structures incorporating AMs. In all models, the RNN calculates the encoded values of the hidden state sequence, denoted as $\boldsymbol{h}_1, \ldots, \boldsymbol{h}_{n_{cy}}$, for each cycle. The model depicted in Fig. 5(a), referred to as the RNN + TA + 1D CNN model, uses these encoded values to calculate TA scores of all hidden states in the sequence. The representative context vector for each cycle is then derived as a linear sum, weighted by the TA scores, effectively reducing memory requirements by assigning a shallow linear layer before 1D CNN. Inspection of the TA scores reveals crucial time steps with significant implications for knee-onset prediction. A limitation of this model, however, is its inability to identify the critical cycles that could be involved in the input reduction process.

### 3.3.2. RNN + CA + 1D CNN

To capture relevant features from inter-cycle trends, i.e., cycle-to-cycle changes in the profiles, we introduce *cyclic attention* (CA), a combination of SA and MHA. As previously discussed, SA is applied to the context vectors of all the cycles to assess similarities between cycles. Leveraging the key-query relationships identified by SA, crucial cycles for knee-onset regression can be discerned. Subsequently, MHA is applied to measure similarities between context vectors from various viewpoints.

The second model, denoted as the RNN + CA + 1D CNN model and illustrated in Fig. 5(b), integrates the baseline RNN + 1D CNN model with CA. In this configuration, the context vector comprises the last hidden state of the RNN for each cycle, and SA is applied to assess the similarities between these context vectors. It assigns elevated weights to significant cycles, generating refined context vectors $\widetilde{\boldsymbol{ct}}_1, \ldots, \widetilde{\boldsymbol{ct}}_{n_{cy}}$ from raw context vectors $\boldsymbol{ct}_1, \ldots, \boldsymbol{ct}_{n_{cy}}$ (outputs from the RNN). The adjustment enables the cycle-to-cycle relationship to be reflected before feeding the context vectors into the CNN. The context vectors are derived from either the last hidden state or the context vector after applying TA to the entire hidden state sequence. An optimal input dimension can be determined by analyzing the CA score from this model.

Compared to the RNN + TA + 1D CNN model, the RNN + CA + 1D CNN model offers the benefit of identifying important cycles. However, it relies on the RNN's last hidden state as the context vector for each cycle, leading to the identification of the crucial cycles while being influenced by the VG. This approach can result in biased information about the main cycle, which may differ from the actual one. To address this issue, we propose a hybrid model, RNN + TA + CA + 1D CNN, combining the strengths of both individual models.

### 3.3.3. RNN + TA + CA + 1D CNN

The RNN + TA + CA + 1D CNN model, depicted in Fig. 5(c), integrates the strengths of both TA and CA. In this model, context vectors are generated using TA scores, akin to the first model, instead of relying solely on the last hidden state of the RNN. This approach, like the RNN + TA + 1D CNN model, reduces memory requirements by utilizing just one hidden state per cycle in the 1D CNN input, effectively mitigating the VG issue by ensuring the context vector encapsulates information from all timesteps within the cycle.

The CA leverages the derived context vector, assembling significant cycles from context vectors that encapsulate all timesteps. This process yields unbiased insights into these cycles, potentially boosting regression performance. It further enables the analysis of input size reduction through the dynamics between queries and keys. Additionally, the application of MHA to SA enhances the model's ability to uncover intricate cycle-to-cycle relationships in varied manners. This enhanced capability reveals critical cycles not discernible through SHA alone, thereby further improving the model's regression performance.

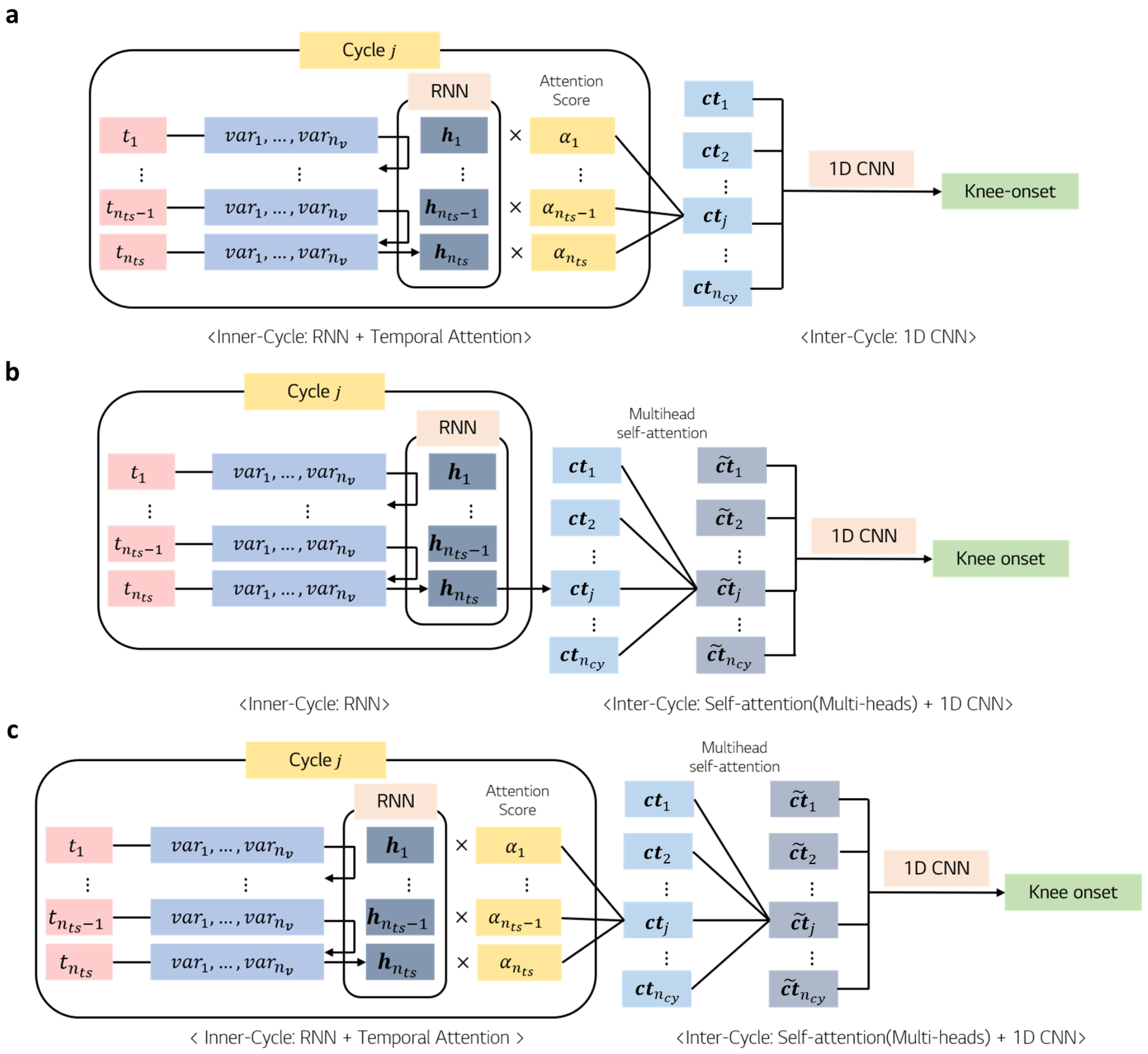

Fig. 5. Attention-based model structures employed in this study. (a) RNN + TA + 1D CNN, (b) RNN + CA + 1D CNN, (c) RNN + TA + CA + 1D CNN.

Table 1. Comparison of characteristics of baseline models and suggested models.

| | Needed Memory | Vanishing Gradient | Explainability Cycles | Timesteps |
|---|---|---|---|---|
| RNN + 1D CNN | Small | O | X | X |
| RNN + 2D CNN | Large | X | X | X |
| RNN + TA + 1D CNN | Small | X | X | X |
| RNN + CA + 1D CNN | Small | O | △ | X |
| RNN + TA + CA + 1D CNN | Small | X | O | O |

# 4. Results and Discussion

In this section, we delve into the analysis of two distinct problems using the Severson dataset. Our primary investigation centers on predicting knee-onset from data collected during the first 100 charging-discharging cycles. Subsequently, we explore the potential to reduce the input dimensionality through the application of CA scores, considering the possibility of leveraging data from a reduced number of initial cycles.

In this study, we utilized three distinct subsets of the Severson dataset, as outlined in our prior research [50]: datasets for charging only, discharging only, and a combined dataset that includes rest periods. Following the methodology established in [50], we divided the data into five unique configurations of training, validation, and test sets, consisting of 80, 20, and 24 cells, respectively. We then calculated the average of the evaluation metrics across these configurations to ensure robust performance assessment. The performance of the various models was evaluated using the root mean square error (RMSE) on the test sets as a measure of regression accuracy. This analysis, including the impact of TA and CA scores, will be further explored in the subsequent section.

## 4.1. 100-cycle scenario

### 4.1.1. Regression performance

To assess the impact of the rest phase on the TA scores, we experimented with two additional datasets (charging only and discharging only) alongside the entire dataset, which includes the rest periods. Table 2 illustrates the performance of the baseline models (i.e., the RNN + 1D CNN and RNN + 2D CNN) and the RNN + TA + 1D CNN model. Across all dataset configurations, the RNN + TA + 1D CNN and RNN + 2D CNN models exhibited lower test losses than the RNN + 1D CNN, primarily by mitigating the VG effect observed in the latter. Among the dataset options, the combined dataset containing data from all charge, rest, and discharge phases is used as inputs for the subsequent experiments.

Table 2. Regression performance of RNN + 1D CNN, RNN + 2D CNN and RNN + TA + 1D CNN. (Charging only, Discharging only, Combined dataset)

| Test loss (RMSE) | Charging only | Discharging only | Combined dataset |
|---|---|---|---|
| RNN + 1D CNN | 149.46 | 98.39 | 93.10 |
| RNN + 2D CNN | 91.94 | **70.68** | **69.25** |
| RNN + TA + 1D CNN | **84.75** | 71.65 | 71.44 |

In Table 3, the performance of the RNN + CA + 1D CNN model is observed to be superior to the RNN + 1D CNN model. This improvement is attributed to considering similarities between cycles in the context vectors constructed by the SA before applying 1D CNN to them. However, owing to the presence of VG, the performance is inferior to models not affected by VG-induced degradation, such as the RNN + 2D CNN or RNN + TA + 1D CNN. While creating a context vector considering the cyclic property is crucial, the limited regression performance is apparent when applying CA solely to the last hidden states, which have lost significant information about transient profiles due to VG. When MHA is applied to RNN + CA + 1D CNN, the prediction performance of 2 heads is better than the SHA but still not as good as the models without the VG problem. With more heads, the performance becomes worse than with one head only. Overall, we can see that VG impacts critically for regression performance which should be resolved in constructing context vectors.

In the subsequent analysis, we evaluate the performance among three models: RNN + TA + CA + 1D CNN, RNN + CA + 1D CNN, and RNN + TA + 1D CNN. For CA, only SHA is considered. The results demonstrate that the model integrating RNN, TA, CA, and 1D CNN surpasses the performance of the other two configurations, highlighting the effective integration of TA and CA. Additionally, in terms of MHA, by using different numbers of attention heads, the findings indicate that MHA consistently outperforms SHA across all configurations, with the utilization of three heads emerging as the most efficient setup in RNN + TA + CA + 1D CNN.

Table 3. Regression performance of baseline and all suggested models for the combined dataset. For models with CA (RNN + CA + 1D CNN, RNN + TA + CA + 1D CNN), MHA regression results are described to show how MHA would benefit knee-onset regression performance.

| Test loss (RMSE) | $n_{he} = 1$ | $n_{he} = 2$ | $n_{he} = 3$ | $n_{he} = 4$ | $n_{he} = 5$ |
|---|---|---|---|---|---|
| RNN + 1D CNN | 93.15 | - | - | - | - |
| RNN + CA + 1D CNN | 79.16 | **76.53** | 79.24 | 84.87 | 85.12 |
| RNN + TA + 1D CNN | 71.44 | - | - | - | - |
| RNN + 2D CNN | 69.25 | - | - | - | - |
| RNN + TA + CA + 1D CNN | **61.71** | 60.71 | **56.23** | 57.64 | 61.03 |

#### 4.1.2. Attention score analysis

Fig. 6 displays the TA scores of cells in each batch across the three datasets (charging only, discharging only, and combined dataset). Fig. 6(a)~(b) presents TA scores for the charging-only and discharging-only datasets, showing minimal difference among the batches. Consequently, when utilizing these datasets, it can be inferred that incorporating TA into the RNN + 1D CNN improves performance solely by alleviating the VG problem. In contrast, Fig. 6(c) (combined dataset) reveals distinct trends in TA scores for Batch 2 compared to Batches 1 and 3. For Batch 2, high TA scores are observed around time step 20 (10 min) for cycles 60-80, corresponding to the rest after charging to 80% SOC. Similar observations are made around time step 100 (50 min), corresponding to the rest period after discharging concludes. These observations underscore the pivotal role of “rest time” in LIB lifespan prediction, justifying the consideration of only the combined dataset is considered for further experiments.

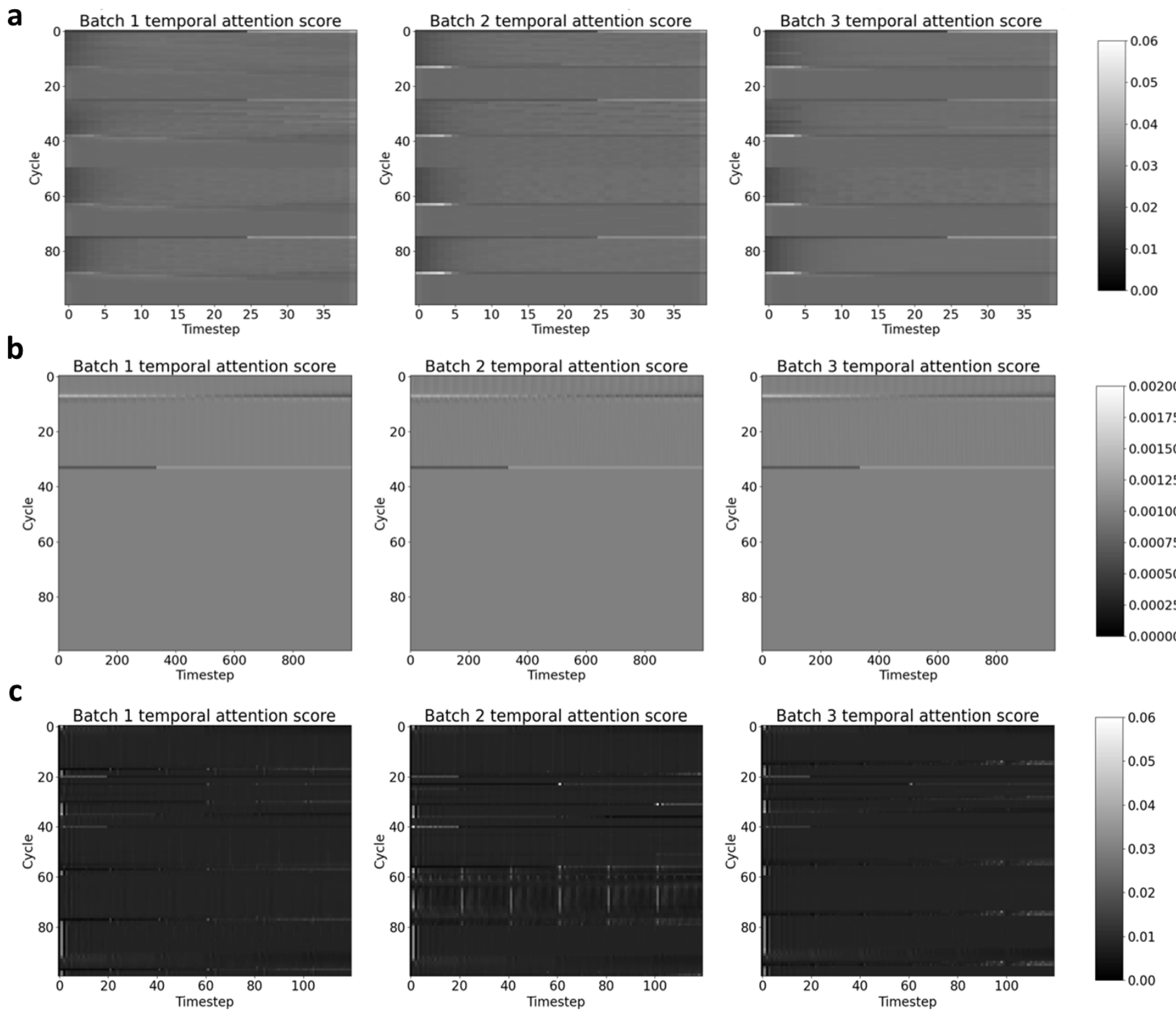

Fig. 6. TA scores of RNN + TA + 1D CNN with different subsets of the Severson dataset. (a) Charging only dataset, (b) Discharging only dataset, (c) Combined dataset. In charging and discharging only, subtle batch-to-batch differences are shown, while in the combined dataset, the TA score is emphasized in sections near the rest phase only in batch 2.

Fig. 7 displays the average CA scores of each batch from the RNN + CA + 1D CNN and RNN + TA + CA + 1D CNN (SHA). In those figures, the x-axis represents the key, and the y-axis represents the query. Batches 1 and 3, sharing similar charging and rest policies, exhibit comparable CA score patterns for both models. In contrast, Batch 2 displays slightly different patterns. Two weaknesses are identified for the RNN + CA + 1D CNN model compared to the RNN + TA + CA + 1D CNN model. Firstly, owing to the vulnerability of RNN + CA + 1D CNN to VG-induced degradation, the context vector of each cycle fails to represent cycle information fully. Consequently, crucial key cycles (indicating high attention scores for nearly all queries) cannot be identified for the RNN + CA + 1D CNN model, hindering the potential reduction of input dimension (Fig. 7(a)).

For RNN + TA + CA + 1D CNN (SHA), the disparity in CA scores between Batches 1, 3 and Batch

2 is more pronounced than in RNN + CA + 1D CNN (Fig. 7(b)). Consequently, the potential to reduce the input dimension based on crucial key cycles is more evident. In Batches 1 and 3, cycles 20-40 and 85-90 emerge as crucial key cycles for regression, with minimal attention given to cycles 60-80. Conversely, in Batch 2, the AM places the highest importance on cycles 60-80. This aligns with the substantial weights assigned to the rest period between cycles 60-80, as depicted in Fig. 6(c). Thus, we conclude that the rest time significantly influences the determination of cycles crucial for the regression. In the next section, based on the different key cycle results for each batch, we will see how much input size can be reduced for two cases: 1) each batch separately and 2) utilizing the entire dataset at once.

The impact of multi-heads for CA is analyzed in Fig. 8, illustrating the CA score patterns from the RNN + TA + CA + 1D CNN model with five heads. Similar to previous observations, the patterns of Batch 2 significantly differ from those of Batches 1 and 3; notably, patterns of heads 2, 4 and 5 exhibit considerable similarity, suggesting redundancy. Consequently, adding more heads beyond 3 harms performance, introducing unnecessary complexity and complicating the training process. Detailed results for various numbers of heads can be found in Fig. S3~S5.

For all batches analyzed, the initial 20 cycles (0-20) were identified as significant by head 1, marking a stark contrast to the SHA scenario where these cycles were overlooked. The introduction of multiple heads enabled the capture of nuanced inter-cycle properties, often obscured by complexity when viewed from singular perspectives. Specifically, we observed that employing multiple heads allows for a considerable reduction in the dimensionality of required input cycles compared to a single-head approach. The efficacy of this dimensionality reduction, facilitated by MHA as opposed to SHA, will be further substantiated in the forthcoming section through a comparative analysis of input size reduction tests using varying input sizes.

Considering that significant cycles range from cycle 60-80 (Batch 2, SHA as CA; Batches 1, 3, head 1 of MHA as CA) 20-40 (Batches 1, 3, SHA as CA; Batch 2, head 1 of MHA as CA; Batches 1 to 3, head 3 of MHA as CA), and 0-20 (Batches 1, 3, head 1 of MHA as CA) in RNN + TA + CA + 1D CNN-based results, 80, 50, 30 are selected as the number of necessary input cycles to predict knee-onset based on the key cycles that contain all high CA-score regions.

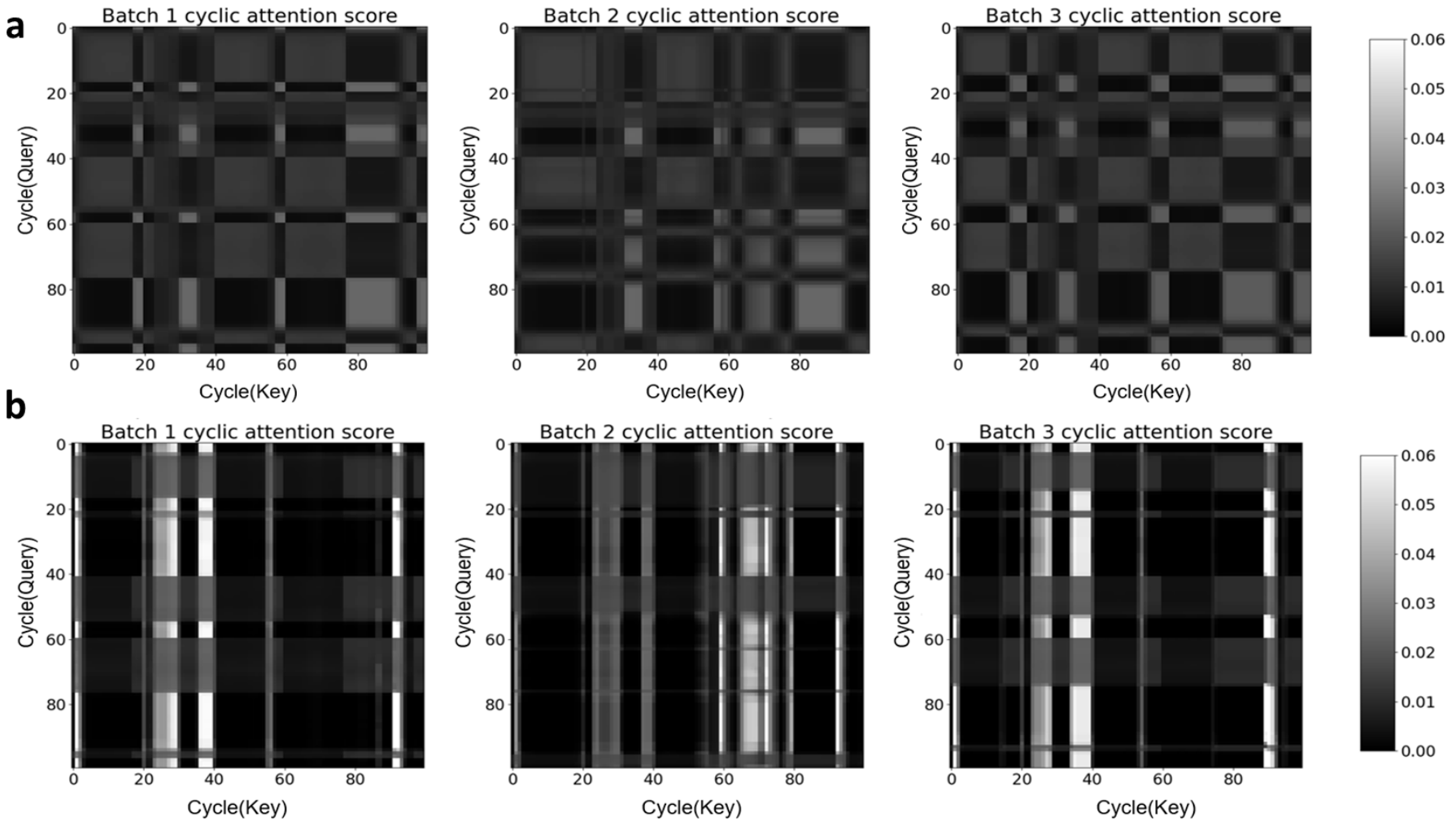

Fig. 7. Average CA scores from RNN + CA + 1D CNN and RNN + TA + CA + 1D CNN (SHA). (a) RNN + CA + 1D CNN, (b) RNN + TA + CA + 1D CNN (SHA). The RNN + CA + 1D CNN model did not single out any cycles as particularly significant across all queries. In contrast, the RNN + TA + CA + 1D CNN (SHA) model identified critical cycles in the ranges of 20-40 for Batches 1/3 and 60-80 for Batch 2. This differentiation is attributed to the model's ability to generate context vectors that account for variations between batches and minimize the VG effect. Consequently, the analysis suggests that the RNN + TA + CA + 1D CNN (SHA) model could potentially reduce the required input cycle size to either 40 or 80 cycles based on the CA score analysis.

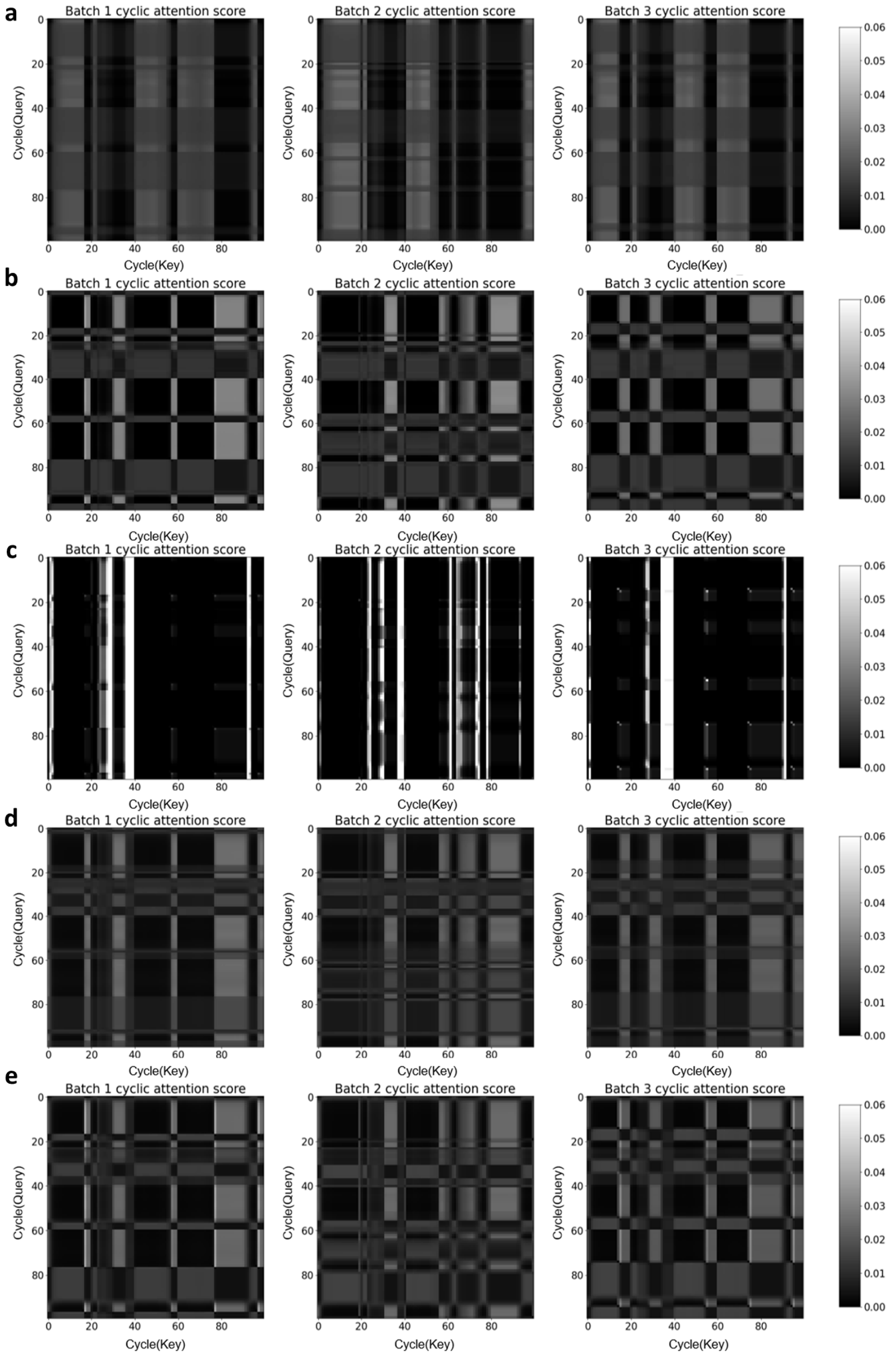

Fig. 8. CA scores from RNN + TA + CA + 1D CNN with five heads. (a) Head 1, (b) Head 2, (c) Head 3, (d) Head 4, (e) Head 5. The first three heads reveal unique patterns of attention, while heads 4 and 5 exhibit patterns closely mirroring those of head 2. Notably, head 1 brings attention to the initial cycles 0-20, a feature not observed with SHA. This highlights the potential for more significant input reduction, as these early cycles are deemed critical by the multi-head attention mechanism but were overlooked in the SHA configuration.

## 4.2. Input reduction scenario

The study showed that crucial information for the knee-onset prediction could be extracted from a subset of the initial 100 cycles. Moreover, important cycles vary for each batch and change with the number of heads used for SA. In this section, we present the results of the input reduction test, where the determination of the number of cycles to include in the input data is based on the SA scores. Two benchmark methods are compared with the suggested method in terms of input reduction possibility and regression performance. For this test, RNN + TA + CA + 1D CNN is trained with SHA and MHA for CA for comparison purposes. The head size for MHA was fixed at 3, as determined from the results of the 100-cycle regression test.

### 4.2.1. Performance comparison with benchmark methods

Table 4 presents the outcomes of input size reduction experiments using SHA and MHA for cyclic attention within the RNN + TA + CA + 1D CNN framework compared with two benchmark methods. Benchmark 1 employs sparse regression using Elastic Net (EN) on Voltage, Current, and Temperature (VIT) data (Directly Measured Health Indicators, DHI) from the cycle 1 to the cycle $n_{cy}$. Benchmark 2 adopts the methodology proposed in [12] , applying EN to leverage features of the "Full" model of [12] (RHIs), identified as optimal for EOL prediction. However, the reliance of this RHI-based approach on data from 100 cycles necessitates knee-onset prediction to be performed solely for $n_{cy}$=100. The findings reveal that the RNN + TA + CA + 1D CNN model significantly surpasses both benchmarks, even when only SHA is employed, with significant margins of 20-140 cycles. This underscores the proposed method's robust regression performance and broad applicability, both in specific experimental conditions (individual batches) and across a diverse spectrum of conditions (all batches). Furthermore, while reducing cycling costs remains crucial, selected features from [12] pose challenges for input size reduction, underscoring their limited field applicability and highlighting the significance of utilizing DHIs where RHIs may be impractical..

Using all batches data together, the input reduction possibilities of RNN + TA + CA + 1D CNN

are discussed in depth. Initially, the input dimensionality was successfully reduced to 50 cycles using SHA for CA, with only minor performance deterioration observed when input sizes were scaled down to 80 and 50 cycles from the standard 100-cycle input. Notably, when the input was further decreased to 30 cycles, a significant increase in error was noted. This escalation in error can be attributed to the removal of critical cycles, particularly evident in Batches 1 and 3, where important cycles with high CA scores span between 20 and 40 cycles, as illustrated in Fig. 8. Reducing the input to 30 cycles in the SHA scenario omits these essential cycles, thereby eliminating crucial degradation-related information beneficial for the 1D CNN's analysis. Consequently, the regression performance suffers considerably with inputs less than 50 cycles, highlighting the importance of including key cycles in the dataset for optimal results.

The application of MHA for CA within the RNN + TA + CA + 1D CNN model consistently outperformed SHA across all scenarios, particularly when the input was limited to 30 cycles. This superior performance of MHA-based CA was evident across various tests, especially with an optimized selection of heads. When reducing the input to 30 cycles, the escalation in test loss was notably lower for MHA compared to SHA. Remarkably, the test error with 30 cycles as input closely mirrored that observed with a full 100-cycle input, underscoring MHA's capability to dramatically reduce input dimensions without compromising performance. This is corroborated by the CA score distribution seen in Fig. 8, where employing three heads enabled the identification of crucial early cycles (0-20 by head 1 and 20-40 by head 3) from diverse angles, thereby minimizing data interpretation bias—a limitation seen with SHA that predominantly emphasized cycles 20-40.

This nuanced approach to selecting cycles for analysis, by integrating CA scores from multiple heads and thereby slightly increasing the model's complexity, resulted in improved regression outcomes even when the input was reduced to 30 cycles. Consequently, we have developed a predictive model that maintains its efficacy with significantly fewer input cycles, as demonstrated in a comparison between using a comprehensive 100-cycle input and a reduced 30-cycle input.

### 4.2.2. Using different batch data independently

Further insights into input reduction across different datasets are gained by examining results for each batch individually, as detailed in Table 4 and illustrated in Fig. S8(b)~(d). This batch-specific analysis, reapplying SHA- and MHA-based CA, reveals nuanced findings. For instance, cells in Batch 2, characterized by an increased IR due to prolonged rest periods, displayed a quicker onset of degradation compared to those in other batches, coupled with a more uniform lifespan distribution. As a result, irrespective of the employed algorithm, Batch 2 consistently reported the lowest average error, indicating its distinct behavior compared to Batches 1 and 3.

Beginning with the SHA-based CA analysis for Batch 2, we observe that reducing the input by 50 cycles leads to an increased average test error. This increase is attributed to the omission of critical cycles between 60-80, as highlighted in Fig. 7(b). Nonetheless, this rise in error is relatively minor and does not substantially affect the overall performance, especially considering the narrower range of test loss in Batch 2 compared to other batches. In contrast, Batches 1 and 3, which exhibit high CA scores between cycles 30 and 40 (as shown in Fig. 7(b)), experience a significant error increase when the input is cut down to 30 cycles. This reduction leads to a marked performance degradation across all batches with a 30-cycle input, with errors increasing approximately by 38 and 9 cycles in Batches 1 and 3, respectively. In contrast, Batch 2 sees a mere three-cycle increment.

When examining the MHA-based CA, it's evident that it consistently outperforms or matches the SHA-based CA across all batches and input sizes. Notably, Batch 1 with MHA-based CA maintains good regression performance with only a four-cycle error increment when using 30 cycles as input, compared to a 100-cycle input scenario. For Batch 3, characterized by complex charging strategies, MHA-based CA significantly lowers the knee-onset prediction error by more than ten cycles in every experiment. Even in Batch 2, MHA-based CA enhances regression accuracy for a 30-cycle input compared to SHA, illustrating MHA's superiority across the board. This is primarily because MHA-based CA identifies critical cycles before cycle 30 as containing essential information, regardless of batch specifics, as depicted in Fig. 8.

This demonstrates that MHA can uncover dominant relationships in datasets with intricate input-output dynamics. It suggests its robust applicability to more complex datasets beyond the Severson dataset, such as real-world scenarios with intricate discharging patterns. With a 30-cycle input, the performance gap between SHA-based and MHA-based CA approaches approximately 15 cycles.

In conclusion, MHA-based CA significantly reduces test loss for all batches by evaluating CA scores from multiple perspectives, enabling a reduction of the input dataset to 30 cycles. This efficiency is underscored in scenarios where all batches are considered, with MHA emphasizing the importance of cycles 0-20 in a 100-cycle scenario across all batches. Notably, MHA demonstrates its effectiveness even in Batch 2, with inherently shorter lifespans and lower average errors, by identifying cycles 0-20 as critical, outperforming SHA in every 30-cycle input test. By facilitating input size reduction through strategic CA score analysis and avoiding the pitfalls of local minima with MHA, this approach offers insights into the minimum number of cycles needed for accurate lifespan prediction under varying conditions with minimal complexity. This method ensures precise knee-onset prediction while accommodating the diverse charging policies and cycle lives of different batteries.

Table 4. Input reduction results for the RNN + TA + CA + 1D CNN model, trained with comprehensive data from all batches as well as individually with data from Batch 1, Batch 2, and Batch 3. This analysis employs MHA utilizing three heads to understand the impact of input dimensionality reduction on model performance.

| Test loss (RMSE) | | 100 cycles | 80 cycles | 50 cycles | 30 cycles |
|---|---|---|---|---|---|
| All batches | Benchmark 1 | 145.77 | 153.32 | 156.58 | 162.43 |
| | Benchmark 2 | 116.55 | - | - | - |
| | RNN + TA + CA + 1D CNN (CA: SHA) | 60.71 | 59.49 | 59.79 | 76.52 |
| | RNN + TA + CA + 1D CNN (CA: MHA) | **56.20** | **57.01** | **55.61** | **58.56** |
| Batch 1 | Benchmark 1 | 133.94 | 138.88 | 140.48 | 140.36 |
| | Benchmark 2 | 91.78 | - | - | - |
| | RNN + TA + CA + 1D CNN (CA: SHA) | 29.42 | 29.73 | 34.16 | 67.36 |
| | RNN + TA + CA + 1D CNN (CA: MHA) | **26.47** | **29.29** | **30.53** | **30.89** |
| Batch 2 | Benchmark 1 | 57.85 | 57.52 | 49.44 | 47.16 |
| | Benchmark 2 | 44.07 | - | - | - |
| | RNN + TA + CA + 1D CNN (CA: SHA) | 17.90 | **15.02** | 19.86 | 21.16 |
| | RNN + TA + CA + 1D CNN (CA: MHA) | **13.10** | 15.80 | **14.14** | **16.15** |
| Batch 3 | Benchmark 1 | 186.04 | 182.05 | 186.16 | 208.75 |
| | Benchmark 2 | 108.77 | - | - | - |
| | RNN + TA + CA + 1D CNN (CA: SHA) | 67.49 | 68.63 | 69.27 | 76.13 |
| | RNN + TA + CA + 1D CNN (CA: MHA) | **59.44** | **57.98** | **59.33** | **61.04** |

# 5. Conclusion

In this study, we harness attention mechanisms, specifically temporal attention (TA) and self-attention (SA), to construct and evaluate three novel data-driven models for predicting the lifespan of LIBs, aiming to bolster interpretability. The models—RNN + TA + 1D CNN, RNN + CA + 1D CNN, and RNN + TA + CA + 1D CNN—successfully illuminate critical timesteps and cycles using a singular shallow layer. These models offer insights into batch-to-batch variances and demonstrate the feasibility of input size reduction based on attention scores, enabling systematic testing of input reduction processes across diverse LIB datasets with preserved performance. The advantages of these models include (1) the use of readily measurable real-time variables like voltage, current, temperature, and capacity as inputs, and (2) robust regression performance even with partial datasets (e.g., charging only or discharging only), facilitating their application under various real-world operational conditions by manufacturers.

The implementation of TA significantly enhanced the baseline RNN + 1D CNN model's performance, primarily by addressing the vanishing gradient (VG) issue. Notably, in Batch 2—characterized by the longest rest durations—high TA scores were allocated to specific rest phases, underlining the critical role of rest time in lifespan prediction and improving model interpretability. The inclusion of rest periods in the dataset revealed pronounced TA score discrepancies, particularly highlighting long rest phases in Batch 2 as pivotal for predicting LIB lifespan and underscoring the importance of incorporating rest phases for enhanced interpretability. The RNN + CA + 1D CNN model discerned similar key factors for Batches 1 and 3 while identifying distinct key factors for Batch 2, reflecting their operational differences.

Among the proposed models, the RNN + TA + CA + 1D CNN exhibited the most outstanding performance, benefiting from the synergistic integration of TA and SA. The CA scores for Batch 2 significantly diverged from those of Batches 1 and 3, attributing the model's superior performance to effective VG issue mitigation and indicating potential for further input dimension reduction. The exploration of multi-head attention (MHA) revealed that while adding more heads initially enhanced performance, exceeding three heads introduced redundancy and increased complexity without yielding additional insights. Through both single-head and multi-head SA approaches, we identified essential key cycles and explored input reduction opportunities, with multi-head configurations offering nuanced insights into further input size minimization.

An input reduction trial utilizing SA scores from the RNN + TA + CA + 1D CNN model demonstrated that, with SHA, the input dimension could be decreased to 50 cycles from 100 with minimal performance degradation. Employing MHA further optimized this reduction to 30 cycles across all scenarios, particularly benefiting Batches 1 and 3. This systematic input reduction methodology, informed by crucial cycle identification from a 100-cycle baseline, enables interpretable and applicable reductions across various datasets, significantly reducing experimental costs associated with LIB lifespan prediction, especially in datasets featuring complex operational strategies.

In conclusion, we propose algorithms that can succinctly extract key temporal and cyclic features via attention mechanisms, offering new insights into LIB cycling datasets previously unattainable with existing methods. These insights, validated against electrochemical background knowledge, facilitate experimental cost reductions. Notably, MHA's potential to discern complex correlations within input sequences was demonstrated, showing superior performance in all evaluated scenarios with just 30 cycles required as input size, thus highlighting its efficacy in battery degradation analysis.

In this research, we examined experimental data obtained from meticulously conducted cycling experiments, which usually persisted until the battery's lifespan concluded. However, the practical aims extend beyond analyzing end-of-life data. Objectives might include the early detection of defects during the battery's formation and testing phases at manufacturing sites, as well as providing quantitative guidance to users on the impact of their charging and discharging habits on battery health, using data derived from actual use. The core methodology we developed is versatile and can be adapted to these situations, laying the groundwork for future research initiatives focused on applying our approach to the early detection of defects and the real-time assessment of battery health using operational data.

## CRediT authorship contribution statement

Jaewook Lee: Conceptualization, Methodology, Investigation, Writing – original draft, Software, Visualization. Seongmin Heo: Writing – review & editing, Supervision. Jay H. Lee: Conceptualization, Writing – review & editing, Supervision.

## Data and Code availability

Severson dataset can be accessed at https://data.matr.io/1/projects/5c48dd2bc625d700019f3204. The codes used in this work based on the Severson dataset, implementing RNN + TA + 1D CNN, RNN + CA + 1D CNN, and RNN + TA + CA + 1D CNN with SHA and MHA for CA, plotting temporal and cyclic attention scores, and the input reduction tests based on the cyclic attention scores, can be found https://github.com/Jaewook-L/Temporal_Cyclic-attention.

# Supplementary Information for
# Enhancing Data Efficiency and Feature Identification for Lithium-Ion Battery Lifespan Prediction by Deciphering Interpretation of Temporal Patterns and Cyclic Variability Using Attention-Based Models

**Jaewook Lee[a], Seongmin Heo[a], Jay H. Lee[a, b, *]**

[a] *Department of Chemical and Biomolecular Engineering, Korea Advanced Institute of Science and Technology, 291 Daehak-ro, Yuseong-gu, Daejeon 34141, Republic of Korea*

[b] *Mork Family Department of Chemical Engineering and Materials Science, University of Southern California, 3651 Watt Way, Los Angeles, CA 90089, United States*

[*] *Corresponding author. E-mail address: jlee4140@usc.edu (J.H. Lee).*

## Supplementary Figures

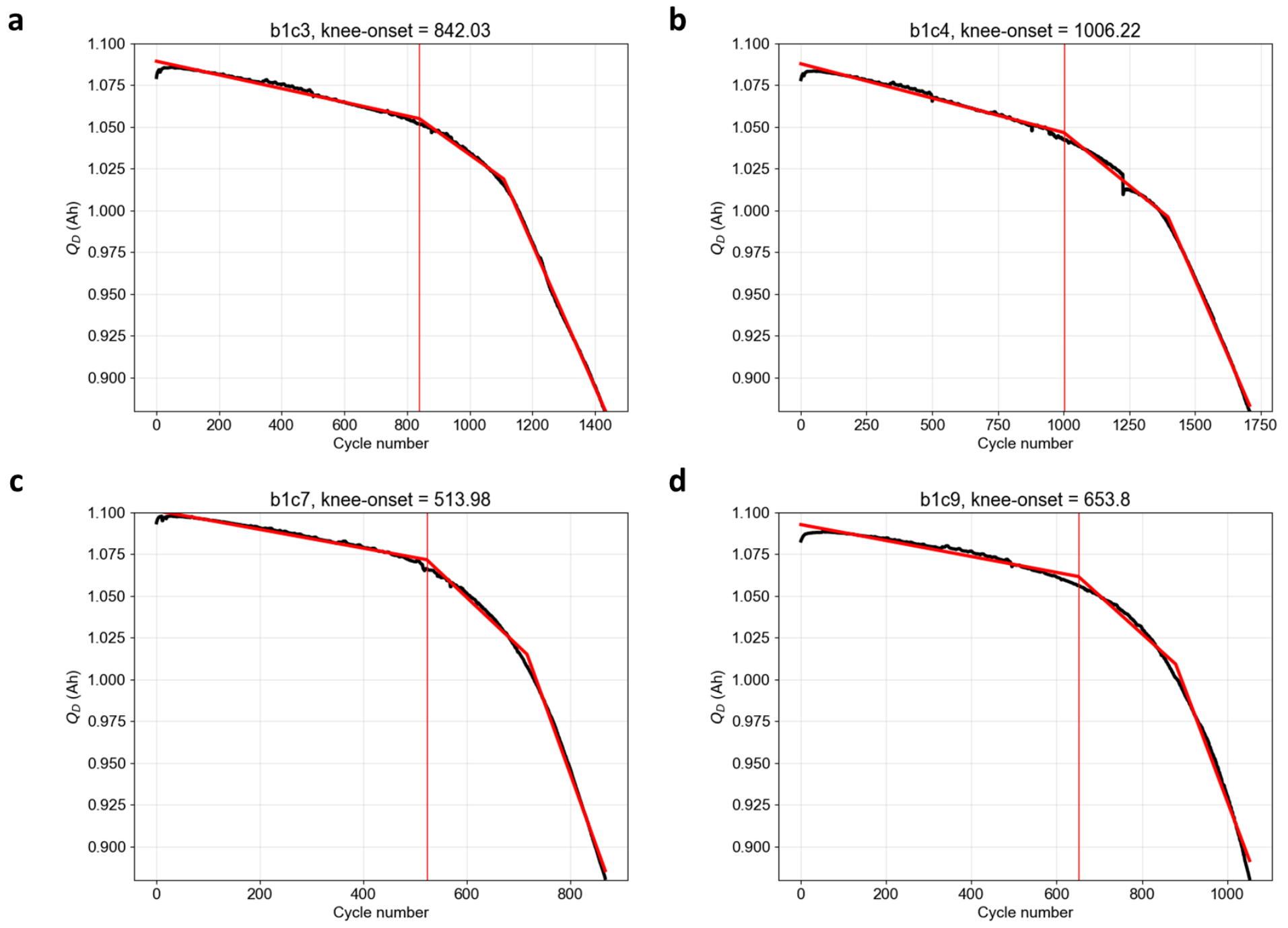

Fig. S1. Knee-onset of some selected cells. (a) b1c3. (b) b1c4, (c) b1c7, (d) b1c9.

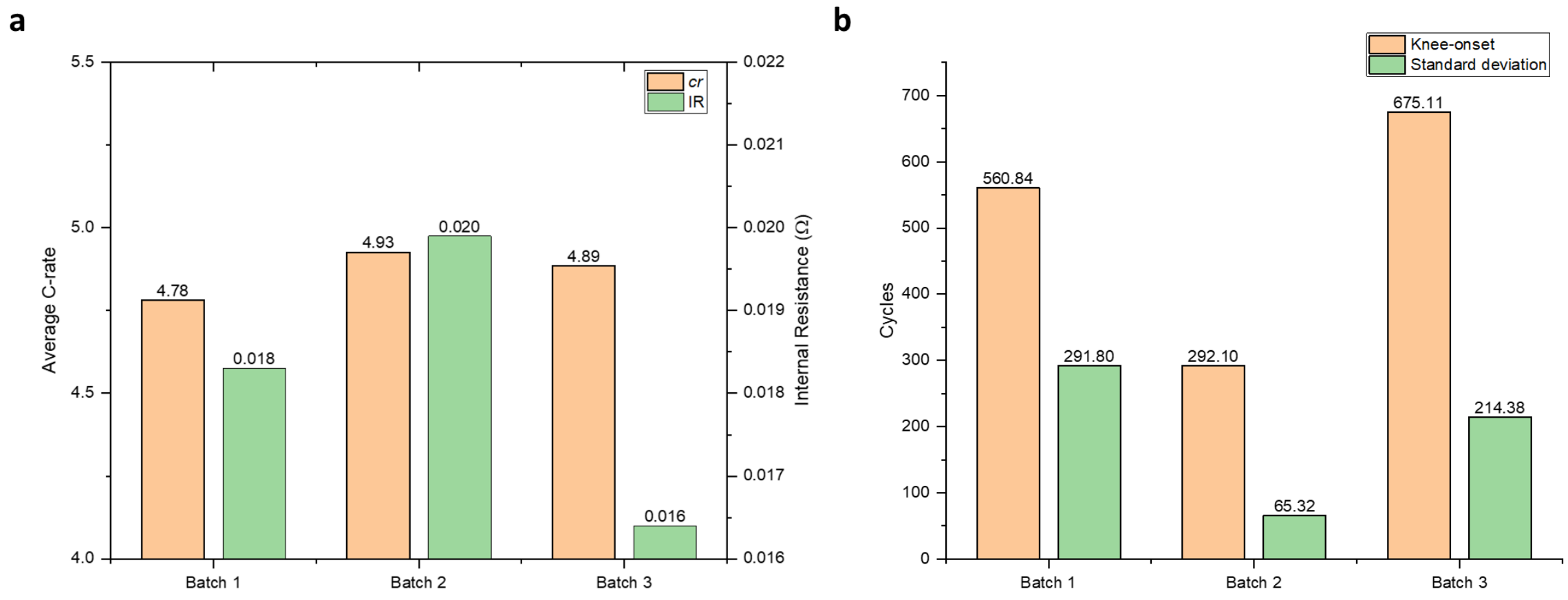

Fig. S2. Different operating conditions and lifespans for each batch. (a) Average C-rate, IR (last 100 cycles) for each batch. (b) Average and standard deviation of knee-onset for each batch.

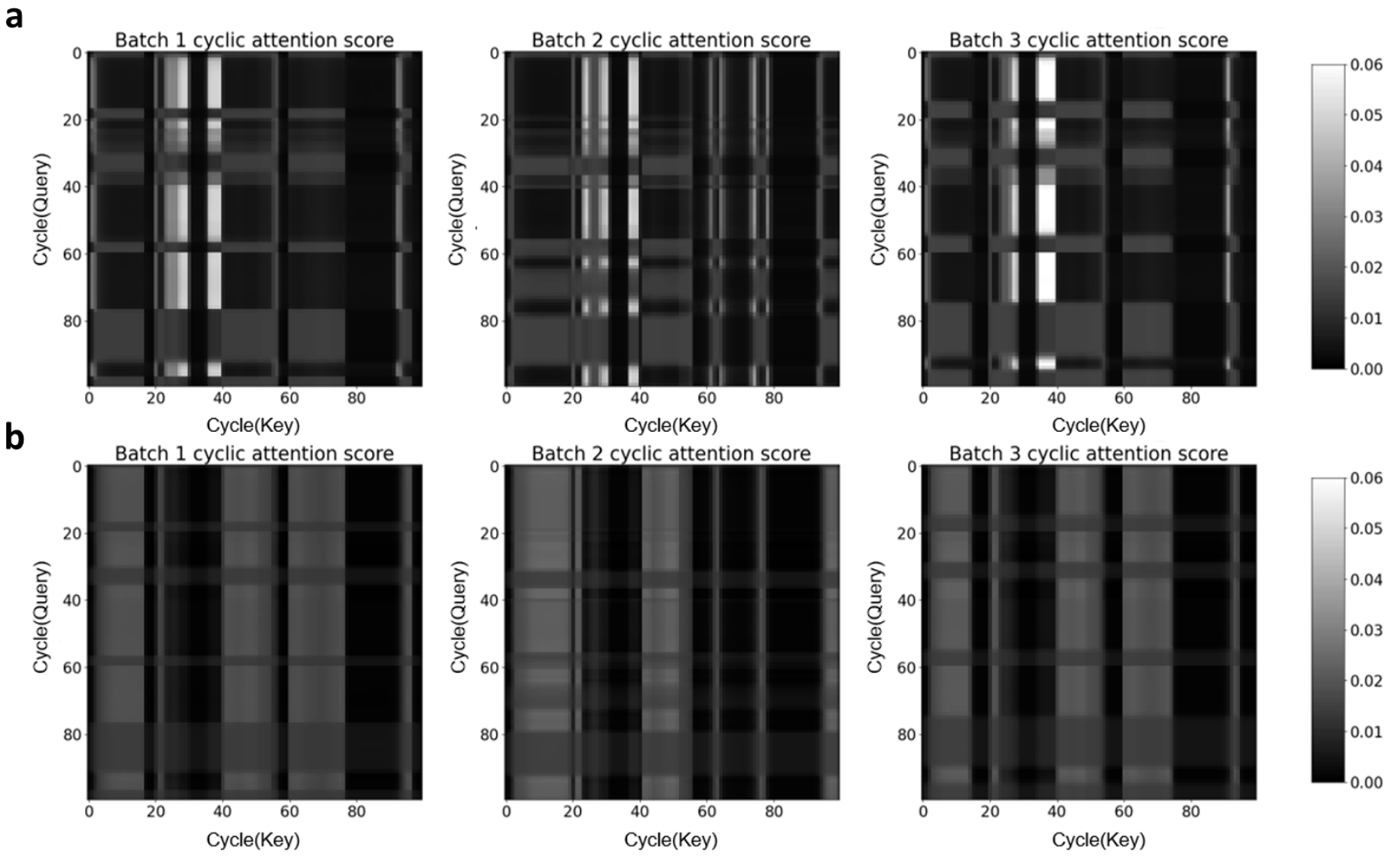

Fig. S3. CA scores from RNN + TA + CA + 1D CNN for 100 cycles as input with two heads. (a) Head 1, (b) Head 2.

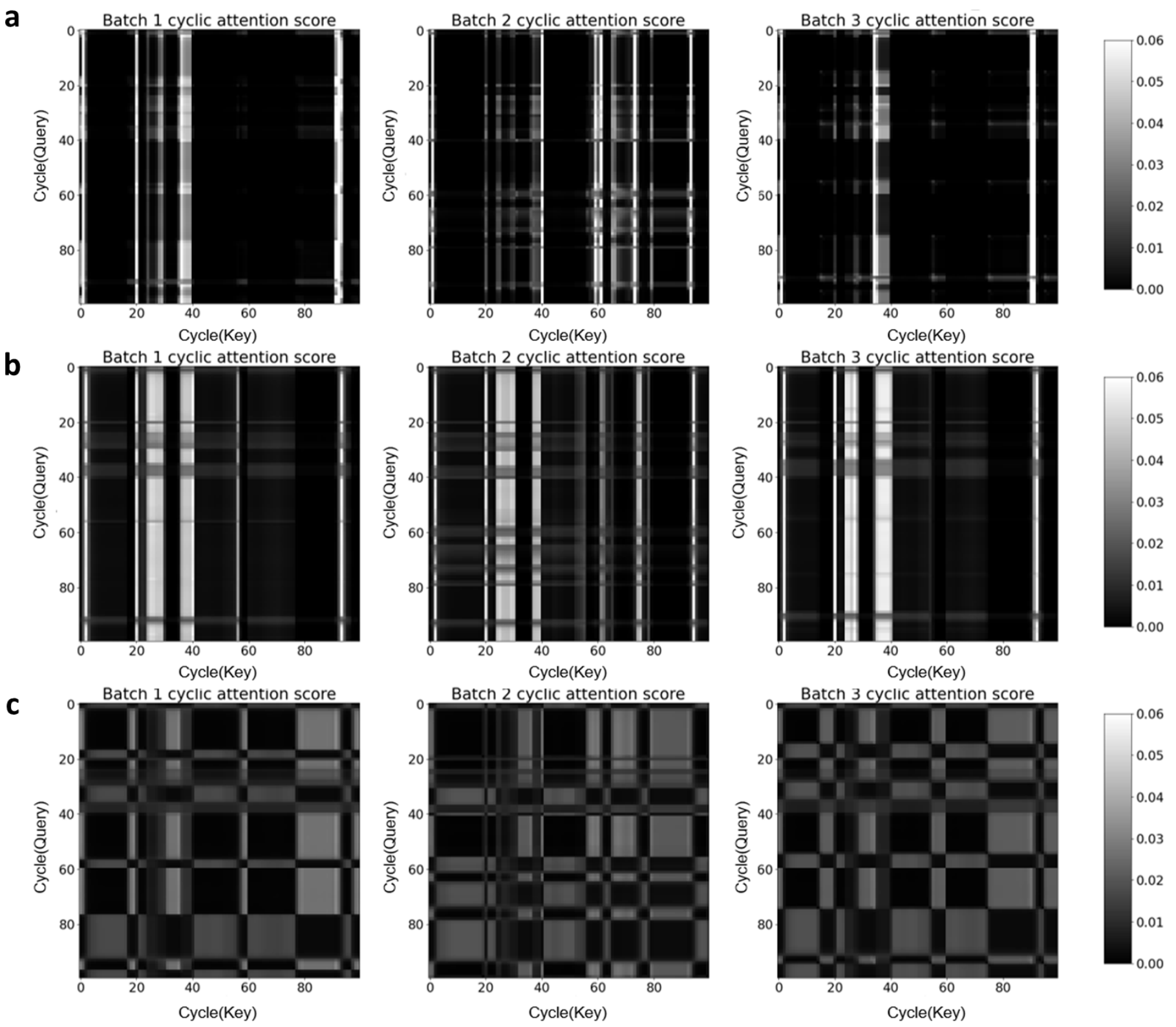

Fig. S4. CA Scores from RNN + TA + CA + 1D CNN for 100 cycles as input with three heads. (a) Head 1, (b) Head 2, (c) Head 3.

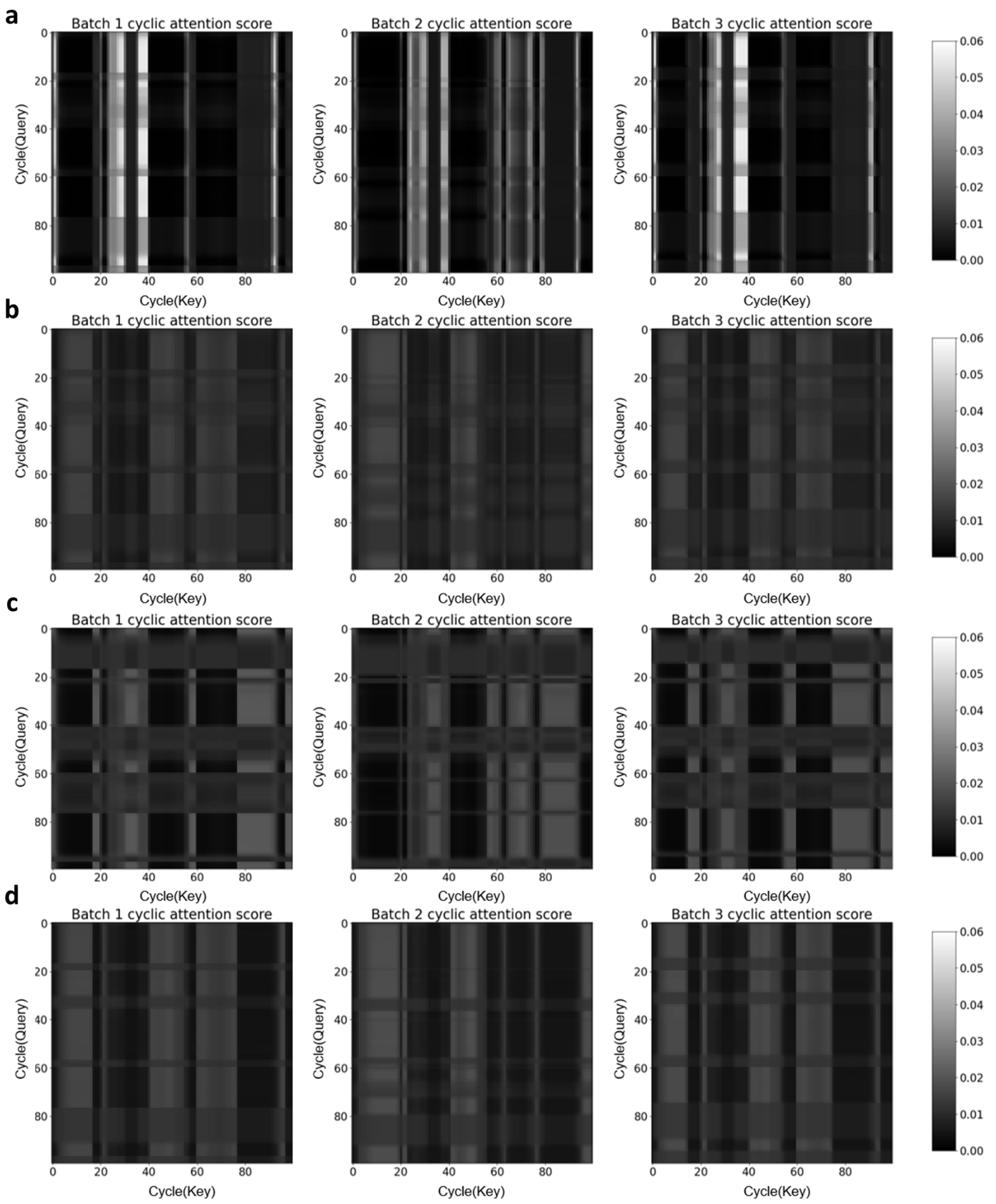

Fig. S5. CA scores from RNN + TA + CA + 1D CNN for 100 cycles as input with four heads. (a) Head 1, (b) Head 2, (c) Head 3, (d) Head 4.

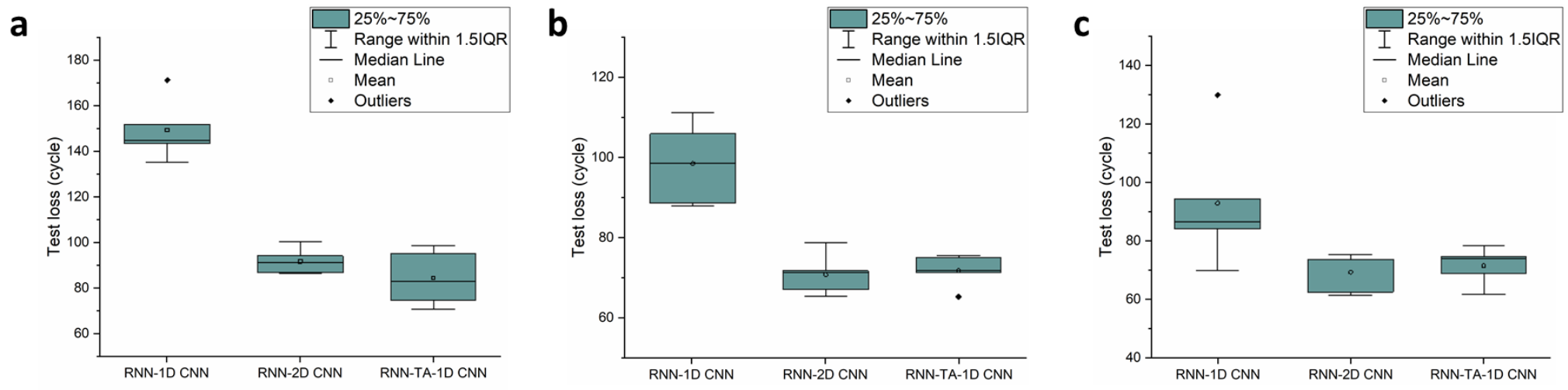

Fig. S6. Losses with error bars for RNN + 1D CNN, RNN + 2D CNN, and RNN + TA + 1D CNN with best hyperparameter values when three different datasets are used with five random seeds. (a) Charging data only, (b) Discharging data only, (c) Combined dataset.

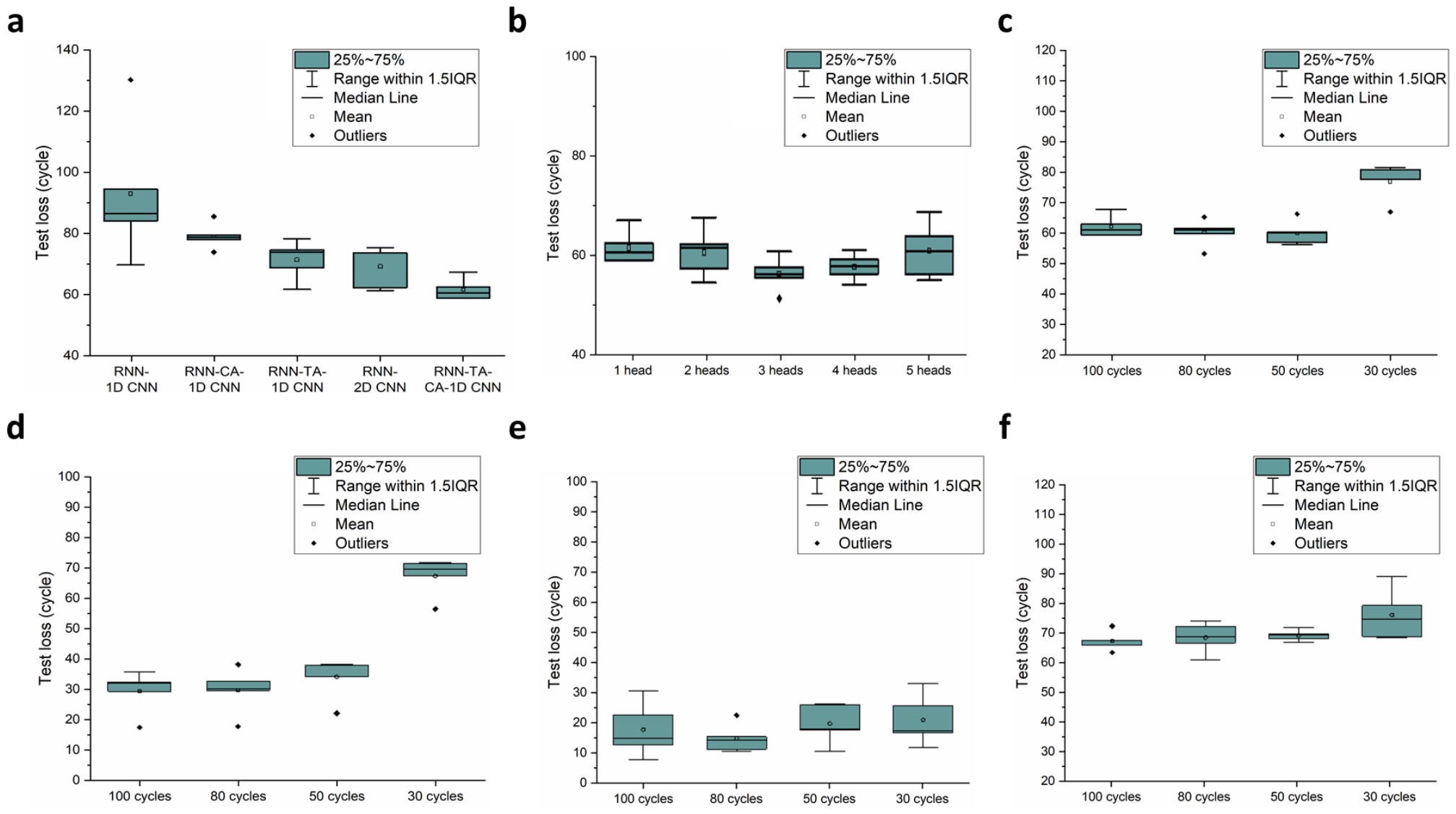

Fig. S7. Test losses with error bars for experiments with RNN + TA + CA + 1D CNN (SHA for CA). (a) Comparison with RNN + 1D CNN, RNN + CA + 1D CNN, RNN + TA + 1D CNN, and RNN + 2D CNN, with best hyperparameter values when the combined dataset is used as an input with five random seeds (SHA is used for CA), (b) Impact of number of heads for CA, (c) Input reductions test results trained with all batches, (d) Input reductions test results trained with Batch 1, (e) Input reductions test results trained with Batch 2, (f) Input reductions test results trained with Batch 3.

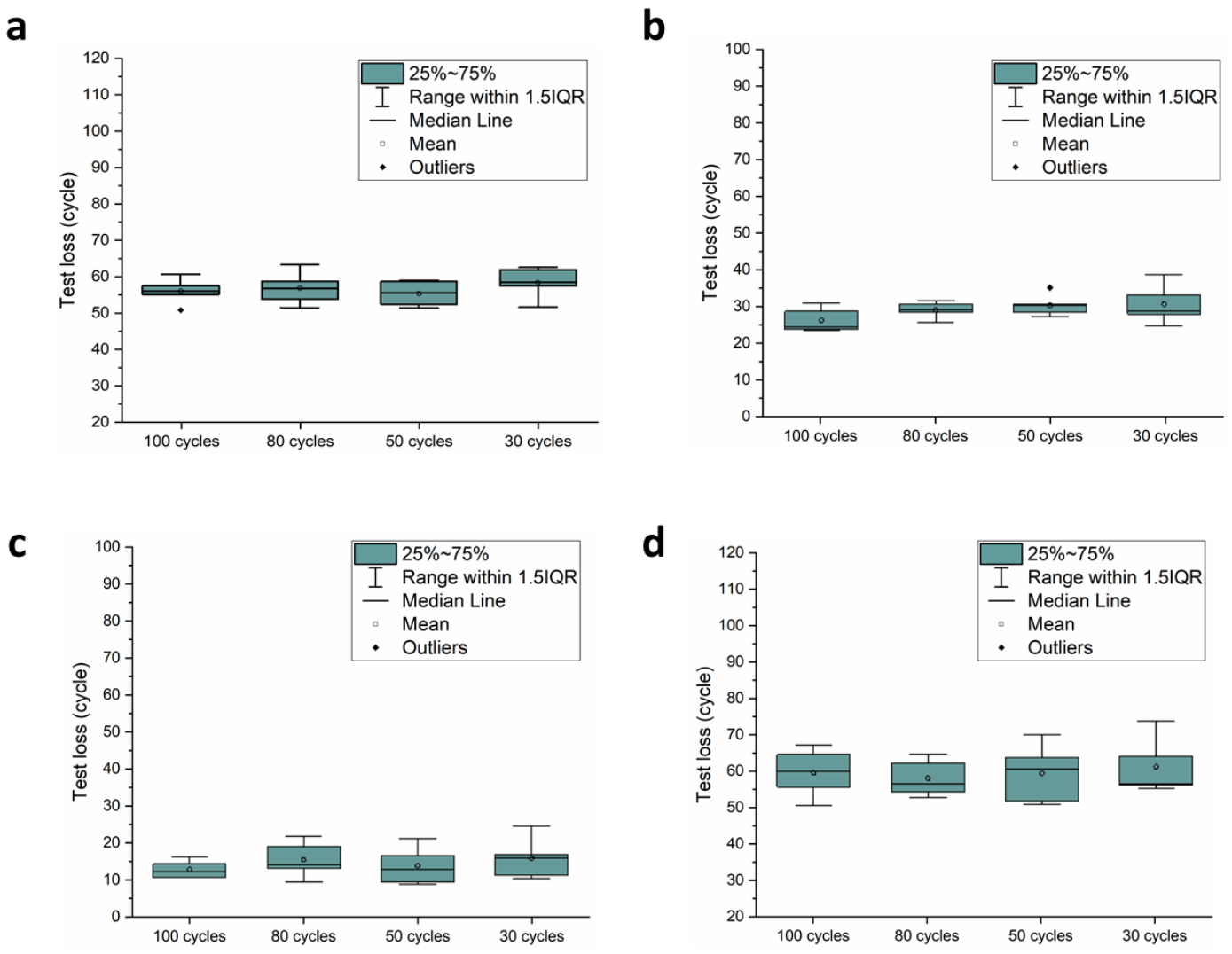

Fig. S8. Error bars of input reduction tests of RNN + TA + CA + 1D CNN (MHA for CA) trained with data from different batches. (a) Input reductions test results trained with all batches, (b) Input reductions test results trained with Batch 1, (c) Input reductions test results trained with Batch 2, (d) Input reductions test results trained with Batch 3.

## Supplementary Tables

Table S1. Cell indexes, Charging policy, Calculated knee-onset. As listed in the reference[1], 'Charging policy' refers to the C-rates between 0% and 80% of SOC and is formatted as "C1(Q1%)-C2", where C1 and C2 represent the C-rates in the first and second phases, respectively, and Q1 represents the SOC where second phase starts.

| index | policy | knee-onset |
|---|---|---|
| b1c0 | 3.6C(80%)-3.6C | 1184.34 |
| b1c1 | 3.6C(80%)-3.6C | 1548.95 |
| b1c2 | 3.6C(80%)-3.6C | 1616.84 |
| b1c3 | 4C(80%)-4C | 842.03 |
| b1c4 | 4C(80%)-4C | 1006.22 |
| b1c5 | 4.4C(80%)-4.4C | 636.07 |
| b1c6 | 4.8C(80%)-4.8C | 435.91 |
| b1c7 | 4.8C(80%)-4.8C | 513.98 |
| b1c9 | 5.4C(40%)-3.6C | 653.80 |
| b1c11 | 5.4C(50%)-3C | 472.27 |
| b1c14 | 5.4C(60%)-3C | 469.17 |
| b1c15 | 5.4C(60%)-3C | 442.19 |
| b1c16 | 5.4C(60%)-3.6C | 508.01 |
| b1c17 | 5.4C(60%)-3.6C | 481.26 |
| b1c18 | 5.4C(70%)-3C | 534.55 |
| b1c19 | 5.4C(70%)-3C | 462.04 |
| b1c20 | 5.4C(80%)-5.4C | 156.01 |
| b1c21 | 5.4C(80%)-5.4C | 338.07 |
| b1c23 | 6C(30%)-3.6C | 584.05 |
| b1c24 | 6C(40%)-3C | 606.17 |
| b1c25 | 6C(40%)-3C | 477.18 |
| b1c26 | 6C(40%)-3.6C | 519.05 |
| b1c27 | 6C(40%)-3.6C | 517.98 |
| b1c28 | 6C(50%)-3C | 464.14 |
| b1c29 | 6C(50%)-3C | 512.14 |
| b1c30 | 6C(50%)-3.6C | 461.56 |
| b1c31 | 6C(50%)-3.6C | 535.01 |
| b1c32 | 6C(60%)-3C | 445.09 |
| b1c33 | 6C(60%)-3C | 454.85 |
| b1c34 | 7C(30%)-3.6C | 475.05 |
| b1c35 | 7C(30%)-3.6C | 428.85 |
| b1c36 | 7C(40%)-3C | 432.73 |
| b1c37 | 7C(40%)-3C | 396.73 |
| b1c38 | 7C(40%)-3.6C | 358.15 |
| b1c39 | 7C(40%)-3.6C | 368.04 |
| b1c40 | 8C(15%)-3.6C | 556.75 |
| b1c41 | 8C(15%)-3.6C | 608.28 |
| b1c42 | 8C(25%)-3.6C | 399.24 |
| b1c43 | 8C(25%)-3.6C | 395.03 |
| b1c44 | 8C(35%)-3.6C | 347.49 |
| b1c45 | 8C(35%)-3.6C | 349.14 |
| b2c0 | 1C(4%)-6C | 123.13 |
| b2c1 | 2C(10%)-6C | 45.09 |
| b2c2 | 2C(2%)-5C | 265.09 |
| b2c3 | 2C(7%)-5.5C | 124.18 |
| b2c4 | 3.6C(22%)-5.5C | 291.93 |
| b2c5 | 3.6C(2%)-4.85C | 299.01 |
| b2c6 | 3.6C(30%)-6C | 399.54 |
| b2c10 | 3.6C(9%)-5C | 354.97 |
| b2c11 | 4C(13%)-5C | 297.60 |
| b2c12 | 4C(31%)-5 | 252.00 |
| b2c13 | 4C(40%)-6C | 316.00 |
| b2c14 | 4C(4%)-4.85C | 301.11 |
| b2c17 | 4.4C(24%)-5C | 308.17 |
| b2c18 | 4.4C(47%)-5.5C | 311.14 |
| b2c19 | 4.4C(55%)-6C | 344.84 |
| b2c20 | 4.4C(8%)-4.85C | 306.85 |
| b2c21 | 4.65C(19%)-4.85C | 298.32 |
| b2c22 | 4.65C(44%)-5C | 318.12 |
| b2c23 | 4.65C(69%)-6C | 336.69 |
| b2c24 | 4.8C(80%)-4.8C | 304.05 |
| b2c25 | 4.8C(80%)-4.8C | 286.11 |
| b2c26 | 4.8C(80%)-4.8C | 285.25 |
| b2c27 | 4.9C(27%)-4.75C | 279.13 |
| b2c28 | 4.9C(61%)-4.5C | 321.98 |
| b2c29 | 4.9C(69%)-4.25C | 311.29 |
| b2c30 | 5.2C(10%)-4.75C | 293.22 |
| b2c31 | 5.2C(37%)-4.5C | 304.48 |
| b2c32 | 5.2C(50%)-4.25C | 302.25 |
| b2c33 | 5.2C(58%)-4C | 329.1 |
| b2c34 | 5.2C(66%)-3.5C | 320.24 |
| b2c35 | 5.2C(71%)-3C | 299.09 |
| b2c36 | 5.6C(25%)-4.5C | 333.38 |
| b2c37 | 5.6C(38%)-4.25C | 307.58 |
| b2c38 | 5.6C(47%)-4C | 300.13 |
| b2c39 | 5.6C(58%)-3.5C | 289.04 |
| b2c40 | 5.6C(5%)-4.75C | 289.24 |
| b2c41 | 5.6C(65%)-3C | 270.46 |
| b2c42 | 6C(20%)-4.5C | 283.03 |
| b2c43 | 6C(31%)-4.25C | 272.15 |
| b2c44 | 6C(40%)-4C | 247.00 |
| b2c45 | 6C(4%)-4.75C | 307.27 |
| b2c46 | 6C(52%)-3.5C | 276.08 |
| b2c47 | 6C(60%)-3C | 456.22 |
| b3c0 | 5C(67%)-4C | 715.95 |
| b3c1 | 5.3C(54%)-4C | 740.77 |
| b3c3 | 5.6C(36%)-4.3C | 757.28 |
| b3c4 | 5.6C(19%)-4.6C | 676.78 |
| b3c5 | 5.6C(36%)-4.3C | 527.11 |
| b3c6 | 3.7C(31%)-5.9C | 433.40 |
| b3c7 | 4.8C(80%)-4.8C | 1283.97 |
| b3c8 | 5C(67%)-4C | 569.00 |
| b3c9 | 5.3C(54%)-4C | 672.26 |
| b3c10 | 4.8C(80%)-4.8C | 666.29 |
| b3c11 | 5.6C(19%)-4.6C | 495.04 |
| b3c12 | 5.6C(36%)-4.3C | 598.09 |
| b3c13 | 5.6C(19%)-4.6C | 496.91 |
| b3c14 | 5.6C(36%)-4.3C | 530.77 |
| b3c15 | 5.9C(15%)-4.6C | 529.99 |
| b3c16 | 4.8C(80%)-4.8C | 968.07 |
| b3c17 | 5.3C(54%)-4C | 837.13 |
| b3c18 | 5.6C(19%)-4.6C | 746.16 |
| b3c19 | 5.6C(36%)-4.3C | 762.07 |
| b3c20 | 5C(67%)-4C | 549.19 |
| b3c21 | 3.7C(31%)-5.9C | 537.88 |
| b3c22 | 5.9C(60%)-3.1C | 702.22 |
| b3c24 | 5C(67%)-4C | 533.15 |
| b3c25 | 5.3C(54%)-4C | 682.11 |
| b3c26 | 5.6C(19%)-4.6C | 647.27 |
| b3c27 | 5.6C(36%)-4.3C | 552.69 |
| b3c28 | 3.7C(31%)-5.9C | 339.59 |
| b3c29 | 5.9C(15%)-4.6C | 525.22 |
| b3c30 | 5.3C(54%)-4C | 638.72 |
| b3c31 | 5.9C(60%)-3.1C | 472.02 |
| b3c33 | 5C(67%)-4C | 931.00 |
| b3c34 | 5.3C(54%)-4C | 679.18 |
| b3c35 | 5.6C(19%)-4.6C | 665.08 |
| b3c36 | 5.6C(36%)-4.3C | 666.66 |
| b3c38 | 5C(67%)-4C | 1449.94 |
| b3c39 | 5.3C(54%)-4C | 841.11 |
| b3c40 | 5.6C(19%)-4.6C | 493.02 |
| b3c41 | 5.6C(36%)-4.3C | 502.75 |
| b3c44 | 5.3C(54%)-4C | 641.14 |
| b3c45 | 4.8C(80%)-4.8C | 947.22 |

Table S2. Tested hyperparameters of the neural networks. ($n_{ep}$: Number of epochs, $f_i$: Number of CNN initial filters, $k_i$: kernel size, $n_p$: Number of CNN pooling layers, $n_{np}$: Number of CNN non-pooling layers, $h_i$: RNN hidden size)

<table>
<tr><th></th><th>Learning rate</th><th>$f_i$</th><th>$k_i$</th><th>$n_p$</th><th>$n_{np}$</th><th>RNN type</th><th>$h_i$ (per head)</th><th>heads (CA)</th></tr>
<tr><td>RNN + 1D CNN</td><td rowspan="2">1e-5, 1e-4, 1e-3, 1e-2</td><td rowspan="2">8</td><td>time = 2~5</td><td rowspan="5">1, 2</td><td rowspan="5">1, 2</td><td rowspan="5">GRU</td><td rowspan="5">3, 5, 7</td><td rowspan="3">-</td></tr>
<tr><td>RNN + 2D CNN</td><td>time = 3, cycle = 2~5</td></tr>
<tr><td>RNN + TA + 1D CNN</td><td rowspan="3">1e-4, 5e-4, 1e-3, 5e-3, 1e-2</td><td rowspan="3">3, 5, 7</td><td rowspan="3">3</td></tr>
<tr><td>RNN + CA + 1D CNN</td><td rowspan="2">1~5</td></tr>
<tr><td>RNN + TA + CA + 1D CNN</td></tr>
</table>

Table S3. Best choices of hyperparameters for the various models with the discharging dataset only.

| | Learning rate | $f_i$ | $k_i$ | $n_p$ | $n_{np}$ | $h_i$ |
|---|---|---|---|---|---|---|
| RNN + 1D CNN | 1e-3 | 8 | 4 | 1 | 1 | 5 |
| RNN + 2D CNN | 1e-4 | 8 | (3,4) | 1 | 1 | 7 |
| RNN + TA + 1D CNN | 1e-2 | 7 | 3 | 2 | 1 | 5 |

Table S4. Best choices of hyperparameters for the various models with the charging dataset only.

| | Learning rate | $f_i$ | $k_i$ | $n_p$ | $n_{np}$ | $h_i$ |
|---|---|---|---|---|---|---|
| RNN + 1D CNN | 1e-4 | 8 | 5 | 1 | 1 | 7 |
| RNN + 2D CNN | 1e-5 | 8 | (3,5) | 2 | 1 | 7 |
| RNN + TA + 1D CNN | 1e-2 | 5 | 3 | 1 | 2 | 5 |

Table S5. Best choices of hyperparameters for the various models with the combined dataset (including charging, discharging, and rest phases) (SHA is used for CA)

| | Learning rate | $f_i$ | $k_i$ | $n_p$ | $n_{np}$ | $h_i$ |
|---|---|---|---|---|---|---|
| RNN + 1D CNN | 1e-2 | 8 | 4 | 2 | 1 | 5 |
| RNN + 2D CNN | 1e-3 | 8 | (3,3) | 2 | 1 | 7 |
| RNN + TA + 1D CNN | 1e-2 | 5 | 3 | 1 | 1 | 7 |
| RNN + CA + 1D CNN | 1e-2 | 7 | 3 | 2 | 1 | 3 |
| RNN + TA + CA + 1D CNN | 1e-2 | 5 | 3 | 2 | 1 | 7 |

Table S6. Best choices of hyperparameters for RNN + TA + CA + 1D CNN when multi-head attention is used for CA. (Input: 100 cycles of the combined dataset, including charging, discharging, and rest phases)

| | Learning rate | $f_i$ | $k_i$ | $n_p$ | $n_{np}$ | $h_i$ per head |
|---|---|---|---|---|---|---|
| 2 heads | 1e-3 | 7 | 3 | 2 | 1 | 7 |
| 3 heads | 1e-2 | 5 | 3 | 1 | 2 | 7 |
| 4 heads | 1e-3 | 5 | 3 | 2 | 2 | 3 |
| 5 heads | 1e-3 | 5 | 3 | 2 | 1 | 7 |

Table S7. Best choices of hyperparameters for RNN + TA + CA + 1D CNN for the input reduction test (All batches)

| | | Learning rate | $f_i$ | $k_i$ | $n_p$ | $n_{np}$ | $h_i$ per head |
|---|---|---|---|---|---|---|---|
| 30 cycles | 1 head | 1e-2 | 3 | 3 | 1 | 1 | 3 |
| | 3 heads | 5e-4 | 5 | 3 | 2 | 1 | 3 |
| 50 cycles | 1 head | 1e-2 | 3 | 3 | 1 | 1 | 3 |
| | 3 heads | 5e-3 | 3 | 3 | 1 | 1 | 3 |
| 80 cycles | 1 head | 1e-4 | 3 | 3 | 1 | 1 | 3 |
| | 3 heads | 1e-4 | 5 | 3 | 2 | 1 | 3 |

Table S8. Best choices of hyperparameters for RNN + TA + CA + 1D CNN for the input reduction test (Batch 1)

| | | Learning rate | $f_i$ | $k_i$ | $n_p$ | $n_{np}$ | $h_i$ per head |
|---|---|---|---|---|---|---|---|
| 30 cycles | 1 head | 1e-2 | 3 | 3 | 1 | 1 | 3 |
| | 3 heads | 5e-3 | 7 | 3 | 2 | 1 | 3 |
| 50 cycles | 1 head | 1e-2 | 3 | 3 | 1 | 1 | 3 |
| | 3 heads | 1e-2 | 7 | 3 | 2 | 1 | 3 |
| 80 cycles | 1 head | 1e-2 | 5 | 3 | 1 | 2 | 3 |
| | 3 heads | 1e-4 | 5 | 3 | 2 | 1 | 5 |

Table S9. Best choices of hyperparameters for RNN + TA + CA + 1D CNN for the input reduction test (Batch 2)

| | | Learning rate | $f_i$ | $k_i$ | $n_p$ | $n_{np}$ | $h_i$ per head |
|---|---|---|---|---|---|---|---|
| 30 cycles | 1 head | 1e-2 | 5 | 3 | 1 | 2 | 3 |
| | 3 heads | 1e-2 | 5 | 3 | 2 | 1 | 7 |
| 50 cycles | 1 head | 1e-2 | 7 | 3 | 1 | 2 | 7 |
| | 3 heads | 1e-2 | 7 | 3 | 1 | 1 | 3 |
| 80 cycles | 1 head | 5e-3 | 5 | 3 | 1 | 1 | 3 |
| | 3 heads | 1e-4 | 5 | 3 | 1 | 2 | 3 |

Table S10. Best choices of hyperparameters for RNN + TA + CA + 1D CNN for the input reduction test (Batch 3)

| | | Learning rate | $f_i$ | $k_i$ | $n_p$ | $n_{np}$ | $h_i$ per head |
|---|---|---|---|---|---|---|---|
| 30 cycles | 1 head | 1e-2 | 5 | 3 | 1 | 1 | 3 |
| | 3 heads | 1e-4 | 3 | 3 | 1 | 1 | 5 |
| 50 cycles | 1 head | 1e-4 | 5 | 3 | 2 | 1 | 7 |
| | 3 heads | 1e-4 | 3 | 3 | 1 | 2 | 3 |
| 80 cycles | 1 head | 1e-4 | 3 | 3 | 2 | 1 | 5 |
| | 3 heads | 1e-2 | 5 | 3 | 1 | 2 | 5 |